\pdfoutput=1
\documentclass[11pt]{article}

\usepackage[preprint]{acl}

\usepackage{times}
\usepackage{latexsym}

\usepackage[T1]{fontenc}
\usepackage[utf8]{inputenc}

\usepackage{microtype}

\usepackage{inconsolata}

\usepackage{graphicx}
\usepackage{amsmath}
\usepackage{amsfonts}
\usepackage{algorithmicx}
\usepackage{algorithm, algpseudocode}
\usepackage{booktabs}
\usepackage{enumitem}
\usepackage{multirow}
\usepackage{tikz}
\usepackage[most]{tcolorbox}
\usepackage{algorithm, algpseudocode}
\usepackage{bbm}
\usepackage[figuresright]{rotating}
\usepackage{longtable}
\usepackage[table]{xcolor}
\usepackage{colortbl}
\usepackage{adjustbox}
\definecolor{mygreen}{RGB}{81, 158, 62}
\definecolor{myblue}{RGB}{121, 176, 223}
\definecolor{myred}{RGB}{221, 127, 114}

\title{Guaranteeing Knowledge Integration with Joint Decoding for Retrieval-Augmented Generation}

\author{
  \textbf{Zhengyi Zhao\textsuperscript{1,5}},
  \textbf{Shubo Zhang\textsuperscript{2}},
  \textbf{Zezhong Wang\textsuperscript{1,5}},
  \textbf{Yuxi Zhang\textsuperscript{2}},
  \textbf{Huimin Wang\textsuperscript{3}},\\
  \textbf{Yutian Zhao\textsuperscript{3}},
  \textbf{Yefeng Zheng\textsuperscript{4}},
  \textbf{Binyang Li\textsuperscript{2}\thanks{Corresponding Author}},
  \textbf{Kam-Fai Wong\textsuperscript{1,5}},
  \textbf{Xian Wu\textsuperscript{3}\footnotemark[1]}
\\
\\
  \textsuperscript{1} The Chinese University of Hong Kong
  \textsuperscript{2} University of International Relations \\
  \textsuperscript{3} Tencent Jarvis Lab
  \textsuperscript{4} Westlake University\\
  \textsuperscript{5} Ministry of Education Key Laboratory of High Confidence Software Technologies, CUHK
\\
  {
  \texttt{\{zyzhao,kfwong\}@se.cuhk.edu.hk},\; \texttt{byli@uir.edu.cn},\; \texttt{kevinxwu@tencent.com}
  }
}

\begin{document}
\maketitle

\begin{abstract}
Retrieval-Augmented Generation (RAG) significantly enhances Large Language Models (LLMs) by providing access to external knowledge. However, current research primarily focuses on retrieval quality, often overlooking the critical ``integration bottleneck'': even when relevant documents are retrieved, LLMs frequently fail to utilize them effectively due to conflicts with their internal parametric knowledge. In this paper, we argue that implicitly resolving this conflict in a single generation pass is suboptimal. We introduce \textsc{GuarantRAG}, a framework that explicitly decouples reasoning from evidence integration. First, we generate an ``Inner-Answer'' based solely on parametric knowledge to capture the model's reasoning flow. Second, to guarantee faithful evidence extraction, we generate a ``Refer-Answer'' using a novel Contrastive DPO objective. This objective treats the parametric Inner-Answer as a negative constraint and the retrieved documents as positive ground truth, forcing the model to suppress internal hallucinations in favor of external evidence during this phase. Finally, rather than naive concatenation or using the DPO trained model directly, we propose a joint decoding mechanism that dynamically fuses the logical coherence of the Inner-Answer with the factual precision of the Refer-Answer at the token level. Experiments on five QA benchmarks demonstrate that \textsc{GuarantRAG} improves accuracy by up to 12.1\% and reduces hallucinations by 16.3\% compared to standard and dynamic RAG baselines.
\end{abstract}

\section{Introduction}

Large Language Models (LLMs) have demonstrated remarkable reasoning capabilities but often struggle with factual accuracy in knowledge-intensive tasks \cite{jiang-etal-2023-active,zeng2024good,xiong2024benchmarking}. Retrieval-Augmented Generation (RAG) addresses this by supplementing the model's parametric memory with external documents \cite{dong2025understand,lin2025explore}. While recent advancements have optimized retrieval precision and dynamic triggering \cite{wu2024repoformer,jeong2024adaptive,yang2024crag}, a critical challenge remains: the effective \textit{integration} of retrieved knowledge.

Current RAG methods typically feed retrieved documents and user queries into the LLM, expecting it to implicitly reconcile the external information with its internal knowledge. However, this approach often leads to two failure modes: (1) \textit{Parametric Override}, where the model ignores retrieved details (e.g., specific statistics) in favor of its generalized internal priors, and (2) \textit{Disjointed Integration}, where external facts are inserted clumsily, disrupting the reasoning flow. Our preliminary experiments on 500 knowledge-intensive queries reveal that 67.3\% of RAG responses suffer from such integration failures, suggesting that the conflict between parametric and non-parametric knowledge is a fundamental bottleneck that cannot be solved by retrieval improvements alone.

To bridge this gap, we introduce \textsc{GuarantRAG}, a framework that transforms knowledge integration from an implicit ``black box'' process into an explicit, multi-stage pipeline. Our core insight is to divide the generation task into two distinct objectives: maintaining reasoning coherence (Parametric) and ensuring factual faithfulness (Non-Parametric).

\textsc{GuarantRAG} operates in three phases. First, we generate an ``Inner-Answer'' using the model's internal knowledge. This captures the LLM's superior reasoning structure and linguistic fluency, albeit with potential hallucinations. Second, we generate a ``Refer-Answer'' dedicated strictly to evidence extraction. A key innovation here is our training strategy: we observe that standard supervised fine-tuning often fails to suppress the model's strong internal priors. To address this, we employ Direct Preference Optimization (DPO) specifically for the Refer-Answer generation. By treating the retrieved documents as the ``chosen'' source and the model's own Inner-Answer as the ``rejected'' negative sample, we explicitly penalize the model for relying on parametric memory when it conflicts with external evidence. This ensures the Refer-Answer is a faithful distillation of the retrieved content. More importantly, we are now aiming to adopt the DPO trained model directly to have the final answer. The model here is only to explicitly distinguish the answer with and without retrieved documents.

Finally, we address the challenge of fusing these two distinct outputs. Simple concatenation of the Inner and Refer answers is insufficient, as it often leads to redundancy or exceeds context windows, causing the model to revert to its priors due to attention dilution. Instead, we propose a Joint Decoding mechanism. This method operates at the token generation level, using the high-level reasoning skeleton of the Inner-Answer while dynamically substituting factual tokens from the Refer-Answer when semantic divergence is detected. This allows \textsc{GuarantRAG} to combine the ``best of both worlds'': the coherent logic of the LLM and the strict accuracy of the retrieval system. Our contributions are as follows:

\begin{figure*}[!t]
    \centering
    \includegraphics[trim={4cm 0 4cm 0},clip,width=\linewidth]{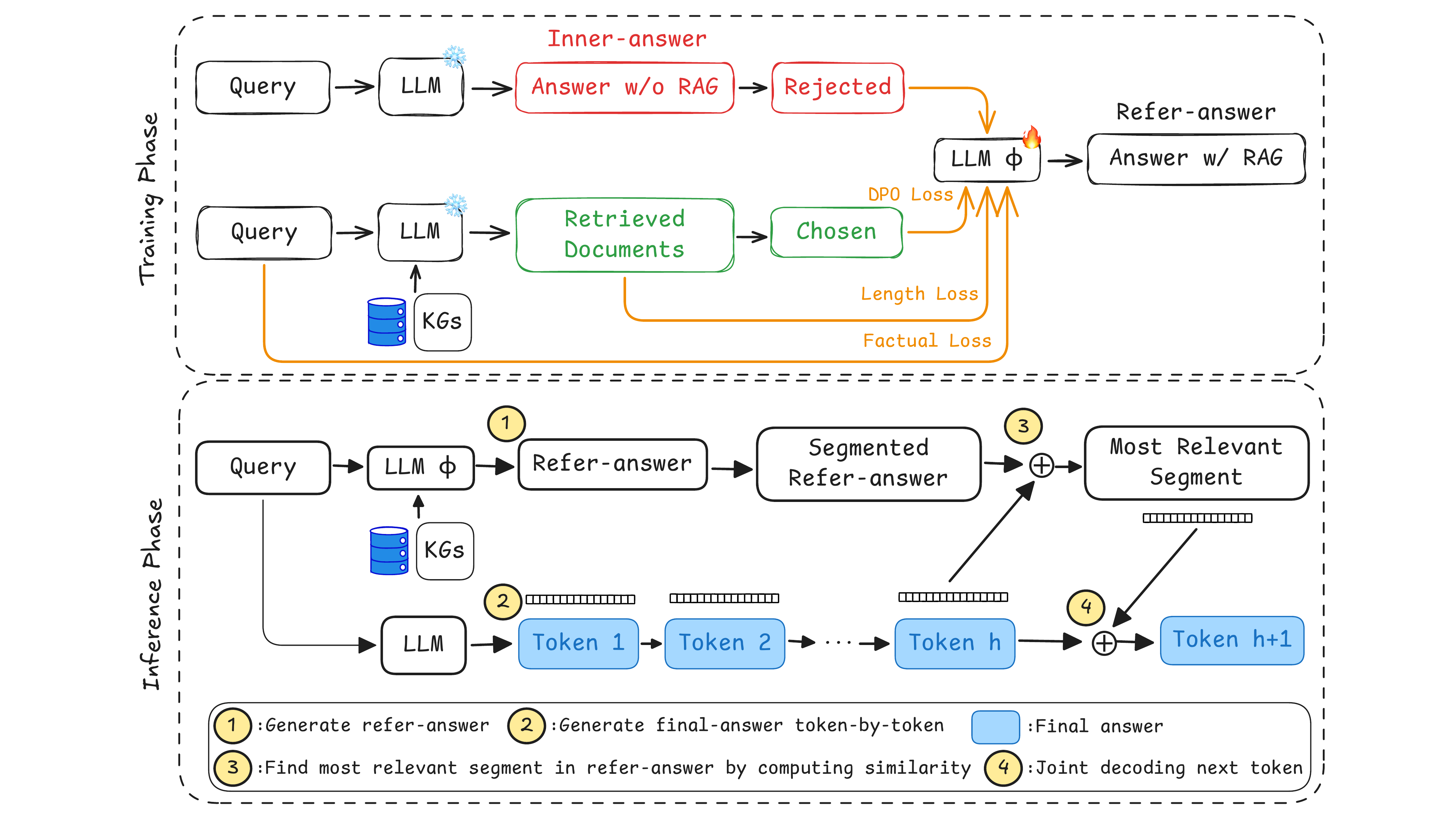}
    \caption{Overview of the \textsc{GuarantRAG} framework. (1) \textbf{Decoupling}: We generate an Inner-Answer (reasoning) and a Refer-Answer (evidence). The Refer-Answer is trained via Contrastive DPO to explicitly prefer retrieved docs over parametric priors. (2) \textbf{Fusion}: A Joint Decoding mechanism dynamically merges the reasoning flow of the Inner-Answer with the factual content of the Refer-Answer during inference.}
    \label{fig:overview}
\end{figure*}

\begin{itemize}[leftmargin=*, itemsep=0pt,parsep=0pt,topsep=0pt,partopsep=0pt]
    \item We identify the conflict between parametric and non-parametric knowledge as the primary cause of RAG integration failure and propose \textsc{GuarantRAG} to explicitly decouple and refuse these knowledge sources.
    \item We introduce a Contrastive DPO training objective that utilizes the model's own Inner-Answer as a negative constraint, guaranteeing that the Refer-Answer is grounded in retrieved evidence rather than internal memory.
    \item We develop a Joint Decoding mechanism that synergizes the reasoning flow of internal knowledge with the factual precision of external documents, outperforming naive concatenation strategies. And extensive experiments show that \textsc{GuarantRAG} improves answer accuracy by up to 12.1\% and reduces hallucinations by 16.3\% across five standard benchmarks.
\end{itemize}

\section{Related Works}

\paragraph{RAG for QA.} RAG has significantly advanced QA capabilities through three primary approaches: (1) static RAG methods \cite{jiang-etal-2023-active,jiang2024piperag,laitenberger2025stronger} that apply a fixed retrieval-then-generate pipeline; (2) adaptive RAG approaches \cite{siriwardhana2023improving,wu2024repoformer,jeong2024adaptive} that dynamically determine when to retrieve based on query characteristics; and (3) iterative RAG systems \cite{wang2024speculative,macdonald2025constructing,hayashi2025iterkey} that refine outputs through multiple retrieval-generation cycles. Despite these advances, existing methods still struggle with effectively integrating parametric and non-parametric knowledge. While recent works have improved retrieval precision \citep{kalra2024hypa,rezaei2025vendi} and query reformulation \cite{wang2024speculative,hayashi2025iterkey}, they typically treat retrieval and generation as sequential rather than interacting processes.

\paragraph{Answer Fusion Techniques.} Answer fusion research has explored three primary integration strategies: (1) context-level fusion \cite{sun2018open,deng2020bridging,wang2023dynamic}, which combines retrieved documents with queries before generation but often leads to information overload; (2) token-level fusion \cite{chen2022grow,mordo2024sponsored}, which integrates information during token generation but frequently sacrifices narrative coherence; and (3) answer-level fusion \cite{rackauckas2024rag,rackauckas2024evaluating,sivasothy2024ragprobe}, which combines separately generated responses but typically operates at coarse granularity. While recent advances in attention mechanisms \cite{li2025attributing,fang2025attentionrag} and ensemble methods \cite{gan2025retrieval,chen2025each} have improved information integration, they still treat retrieved content uniformly without considering its relationship to the model's parametric knowledge. Our \textsc{GuarantRAG} framework introduces a novel segment-level contrastive fusion mechanism that operates at a finer granularity than existing methods, explicitly decomposing answers into semantic segments and performing targeted alignment between complementary knowledge sources.

\section{Methodology}

We propose \textsc{GuarantRAG}, a framework designed to resolve the conflict between parametric memory and external evidence in RAG systems. As illustrated in Figure~\ref{fig:overview}, our approach operates in three stages: (1) a \textit{Knowledge Decision} module that filters unnecessary retrieval; (2) a \textit{Dual-Path Generation} phase that explicitly decouples the model's reasoning flow (Inner-Answer) from factual evidence (Refer-Answer) via contrastive alignment; and (3) a \textit{Joint Decoding} mechanism that dynamically fuses these streams at the token level.

\subsection{Problem Formulation}

Given a knowledge-intensive query $q$ and a set of retrieved documents $D = \{d_1, \ldots, d_k\}$, the goal of a RAG system is to generate a response $y$. Standard methods model this as $P(y|q, D)$ directly. However, we posit that this formulation forces the LLM to implicitly resolve the conflict between its pre-trained parametric knowledge $\mathcal{K}_{\rm param}$ and the non-parametric context $D$, often resulting in hallucinations or disjointed integration.

We reformulate the generation process by decomposing the output into two intermediate latent states: a reasoning-focused \textit{Inner-Answer} $y_{\rm inner}$ derived from $\mathcal{K}_{\rm param}$, and an evidence-focused \textit{Refer-Answer} $y_{\rm ref}$ derived from $D$. The final response $y$ is generated via a joint decoding policy $\pi_{\rm joint}$ that harmonizes these states:
\begin{equation}
    y \sim \pi_{\rm joint}(\cdot | q, y_{\rm inner}, y_{\rm ref})
\end{equation}

\subsection{Phase 1: Knowledge Decision}
To ensure computational efficiency, we first determine if retrieval is necessary. We fine-tune a lightweight estimator based on the base LLM to predict a binary retrieval signal $z \in \{0, 1\}$. The estimator evaluates: (1) \textbf{Temporal Relevance}: is the query outside the training cutoff? and (2) \textbf{Factual Specificity}: does the query require long-tail entity knowledge? If $z=0$, we generate directly. If $z=1$, we proceed to the dual-path framework.

\subsection{Phase 2: Dual-Path Answer Generation}

This phase generates the two intermediate states required for our joint decoding.

\paragraph{Path A: Inner-Answer Generation.}
We first utilize the base model $\pi_\theta$ to generate an inner-answer $y_{\rm inner} \sim \pi_\theta(\cdot|q)$. While potentially factually inaccurate, $y_{\rm inner}$ preserves the model's optimal reasoning structure, linguistic fluency, and instruction-following capabilities. It serves as a ``structural template'' for the final generation.

\paragraph{Path B: Refer-Answer Generation.}
The core challenge in RAG is ensuring the model prioritizes $D$ over $\mathcal{K}_{\rm param}$. To address this, we train a specialized Evidence Model $\pi_\phi$, which initialized from $\pi_\theta$, to generate a \textit{Refer-Answer} $y_{\rm ref}$. 
Unlike standard fine-tuning, we employ a Contrastive DPO objective to explicitly suppress parametric hallucinations. We construct preference pairs where the retrieved document content $D$ is the \textit{chosen} response $y_w$, and the model's own hallucinated $y_{\rm inner}$ is the \textit{rejected} response $y_l$. The objective is:
\begin{equation}
\begin{aligned}
    &\mathcal{L}_{\text{DPO}}(\phi) = \\
    &-\log \sigma \left( \beta \log \frac{\pi_\phi(y_w|x)}{\pi_{\text{ref}}(y_w|x)} - \beta \log \frac{\pi_\phi(y_l|x)}{\pi_{\text{ref}}(y_l|x)} \right)
\end{aligned}
\end{equation}
where $x=(q, D)$. This forces $\pi_\phi$ to treat parametric priors as negative constraints, ensuring $y_{\rm ref}$ is a faithful reflection of external evidence. \textbf{It should be noted that although we regard the inner-answer as reject sample, sometimes the inner-answer is already the correct answer. The purpose here is to use DPO training the model to distinguish the two kinds of answer, instead of adopt this model directly to have the final answer.}

To prevent the degeneration often seen in DPO (e.g., verbatim copying or loss of coherence), we incorporate two auxiliary constraints: \textit{1. Adaptive Length Regularization:} We align the length of the evidence summary with the reasoning path to ensure compatibility during fusion:
\begin{equation}
\mathcal{L}_{\text{len}}(\phi) = \max(0, |y_{\rm ref}| - |y_{\rm inner}|)^2
\end{equation}
\textit{And 2. Factual Relevance Constraint:} To ensure semantic alignment with the query, we maximize the cosine similarity between the embeddings of the query $E(q)$ and the generated refer-answer $E(y_{ref})$:
\begin{equation}
\mathcal{L}_{\text{fact}}(\phi) = 1 - \cos(E(q), E(y_{\rm ref}))
\end{equation}
The final training objective for the Evidence Model is $\mathcal{L} = \mathcal{L}_{\text{DPO}} + \lambda_1 \mathcal{L}_{\text{len}} + \lambda_2 \mathcal{L}_{\text{fact}}$.

\subsection{Phase 3: Joint Decoding with Dynamic Fusion}

The final step is to synthesize the reasoning of $y_{\rm inner}$ with the facts of $y_{\rm ref}$. Naive concatenation of these outputs often fails due to attention dilution. Instead, we propose a Dynamic Joint Decoding mechanism that operates at the hidden-state level.

\paragraph{LLM-based Semantic Segmentation.}
To facilitate precise knowledge injection, we first decompose the evidence-rich refer-answer $y_{\text{ref}}$ into a set of discrete semantic segments $S_{\text{ref}} = \{s_1, \ldots, s_m\}$. Unlike heuristic splitting (e.g., by punctuation) which often leaves multiple distinct facts entangled within complex sentences, we employ a lightweight auxiliary LLM to break the text into \textit{atomic semantic units}. This LLM-driven decomposition is critical for two reasons: (1) \textbf{Granularity}: It ensures that each segment $s_j$ contains exactly one piece of factual evidence, preventing the model from retrieving irrelevant context during fusion; and (2) \textbf{Robustness}: It resolves ambiguous pronoun references within sentences, ensuring that segments remain semantically self-contained for accurate similarity matching during the decoding phase.

\paragraph{Token-Level Fusion.}
We generate the final response $y$ using the base model $\pi_\theta$. At each decoding step $t$, let $\mathbf{h}_t^{\rm gen}$ denote the hidden state of the current token being generated. We simultaneously maintain a context window of the currently generated sentence segment, denoted as $s_{\rm curr}$.
We compute the relevance of the current generation context to the evidence segments:
\begin{equation}
    w_{j} = \text{softmax}\left( \frac{\mathbf{h}(s_{\rm curr})^\top \mathbf{h}(s_j)}{\tau} \right)
\end{equation}
where $\mathbf{h}(\cdot)$ is a sentence encoder. We identify the most relevant evidence segment $s^* = s_{\arg\max(w_j)}$.

To integrate this evidence, we perform a \textit{soft intervention} on the hidden state before the projection to the vocabulary space. The enhanced hidden state $\tilde{\mathbf{h}}_t$ is computed as:
\begin{equation}
    \tilde{\mathbf{h}}_t = \mathbf{h}_t^{\rm gen} + \gamma \cdot \mathbf{h}(s^*)
\end{equation}
where $\gamma$ is a gating factor controlling the strength of external knowledge injection. The final token probability is computed via $P(y_t | y_{<t}) = \text{Softmax}(W_v \tilde{\mathbf{h}}_t)$. This mechanism allows the model to follow the reasoning skeleton of its parametric knowledge $\mathbf{h}_t^{\rm gen}$ while being dynamically ``steered'' toward factual accuracy by the retrieved evidence $\mathbf{h}(s^*)$ whenever the context aligns.

\section{Experiments}

\subsection{Experimental Settings}

\paragraph{Datasets.} We evaluate on five diverse QA benchmarks: Natural Questions (NQ) \cite{kwiatkowski2019natural}, TruthfulQA \cite{lin2022truthfulqa}, Wizard of Wikipedia (WoW) \cite{dinanwizard}, HotpotQA \cite{yang2018hotpotqa}, and ELI5 \cite{fan2019eli5}. These datasets span various knowledge integration scenarios from factoid queries to multi-hop reasoning tasks.

\paragraph{Baselines.} We compare \textsc{GuarantRAG} against both non-RAG and state-of-the-art RAG approaches. The non-RAG baseline uses Qwen3-8B without retrieval augmentation. RAG baselines include Standard RAG, SelfRAG \cite{asai2023self}, RQ-RAG \cite{chan2024rq}, SOLAR \cite{kim2024solar}, DA-RAG \cite{su2025parametric}, FLARE \cite{jiang2023active}, DRAGIN \cite{su2024dragin}, and P-RAG \cite{su2025parametric}. All methods are implemented with Qwen3-8B for fair comparison. For retrieval, we employ BM25 \cite{robertson1994some}, RetroMAE \cite{xiao2022retromae}, SPLADE-v3 \cite{lassance2024splade}, and HyDE \cite{gao2023precise} as different retrievers.

\paragraph{Evaluation Metrics.} We evaluate our approach using metrics across three distinct levels of assessment. At the lexicon level, we employ \textbf{Match}, \textbf{ROUGE-L} , and \textbf{BLEU-4}. At the semantic level, we utilize \textbf{BERTScore} and \textbf{BEM}. For reference-document usage, we introduce three specialized metrics: \textbf{Hallucination Rate (Hal.)} to quantify factually unsupported claims, \textbf{Entity Precision (Ent.)} to measure accurate entity reproduction from references, and \textbf{Structure Coherence (Struc.)} to evaluate logical organization on a 5-point scale through human assessment. Detailed settings and metrics can be found in Appendix~\ref{apd:experimental settings}.

\begin{table*}[!t]
\centering
\small
\adjustbox{max width=\linewidth}{
\begin{tabular}{lccccccccc}
\toprule
\multirow{2}{*}{\textbf{Method}} & \multicolumn{5}{c}{\textbf{Performance Across Datasets}} & \multirow{2}{*}{\textbf{Average}} & \multicolumn{3}{c}{\textbf{Reference Usage}} \\
\cmidrule(lr){2-6} \cmidrule(lr){8-10}
& \textbf{NQ} & \textbf{TruthfulQA} & \textbf{WoW} & \textbf{HotpotQA} & \textbf{ELI5} & & \textbf{Hal.$\downarrow$} & \textbf{Ent.$\uparrow$} & \textbf{Struc.$\uparrow$} \\
\midrule
\multicolumn{10}{l}{\textit{Qwen3-8B Based Methods}} \\
Qwen3-8B & 59.8 & 52.3 & 56.2 & 48.7 & 57.1 & 54.8 & 38.2 & -- & 4.72 \\
\; w/ Standard RAG & 63.4 & 58.1 & 62.7 & 54.3 & 61.5 & 60.0 & 32.6 & 76.8 & 4.15 \\
\; w/ SelfRAG & 67.3 & 61.4 & 66.8 & 59.2 & 67.4 & 64.4 & 28.1 & 81.3 & 4.28 \\
\; w/ RQ-RAG & 68.9 & 62.7 & 68.1 & 61.5 & 68.7 & 65.9 & 26.4 & 82.9 & 4.31 \\
\; w/ SOLAR & 70.4 & 64.2 & 69.5 & 63.1 & 70.2 & 67.5 & 24.7 & 84.5 & 4.37 \\
\; w/ DA-RAG & 66.1 & 60.8 & 65.4 & 57.9 & 66.2 & 63.3 & 29.5 & 80.1 & 4.22 \\
\; w/ FLARE & 69.2 & 63.5 & 68.7 & 62.4 & 69.8 & 66.7 & 25.8 & 83.7 & 4.35 \\
\; w/ DRAGIN & 69.6 & 63.9 & 69.1 & 62.8 & 70.1 & 67.1 & 25.3 & 84.2 & 4.38 \\
\; w/ P-RAG & 71.8 & 65.4 & 71.2 & 64.7 & 72.1 & 69.0 & 22.9 & 86.1 & 4.43 \\
\; w/ \textbf{GuarantRAG} & \textbf{76.4} & \textbf{69.8} & \textbf{75.1} & \textbf{68.9} & \textbf{74.6} & \textbf{74.0} & \textbf{15.7} & \textbf{88.4} & \textbf{4.68} \\
\midrule
\multicolumn{10}{l}{\textit{Qwen3-14B Based Methods}} \\
Qwen3-14B & 62.4 & 54.9 & 58.7 & 51.3 & 59.2 & 57.3 & 35.8 & -- & 4.76 \\
\; w/ Standard RAG & 66.8 & 60.7 & 65.1 & 57.2 & 64.3 & 62.8 & 30.4 & 78.9 & 4.19 \\
\; w/ SelfRAG & 70.1 & 64.2 & 69.3 & 62.5 & 70.1 & 67.2 & 25.9 & 83.7 & 4.32 \\
\; w/ RQ-RAG & 71.9 & 65.8 & 70.9 & 64.3 & 71.8 & 68.9 & 24.1 & 85.2 & 4.35 \\
\; w/ SOLAR & 73.6 & 67.1 & 72.4 & 66.0 & 73.2 & 70.5 & 22.3 & 86.8 & 4.41 \\
\; w/ DA-RAG & 69.3 & 63.5 & 68.2 & 60.8 & 69.0 & 66.2 & 27.2 & 82.4 & 4.26 \\
\; w/ FLARE & 72.4 & 66.3 & 71.5 & 65.1 & 72.7 & 69.6 & 23.5 & 86.1 & 4.39 \\
\; w/ DRAGIN & 72.8 & 66.7 & 72.0 & 65.6 & 73.1 & 70.0 & 23.0 & 86.5 & 4.42 \\
\; w/ P-RAG & 75.2 & 68.9 & 74.3 & 68.1 & 75.8 & 72.5 & 20.4 & 88.7 & 4.47 \\
\; w/ \textbf{GuarantRAG} & \textbf{79.1} & \textbf{72.4} & \textbf{77.8} & \textbf{71.6} & \textbf{78.3} & \textbf{75.8} & \textbf{13.2} & \textbf{90.9} & \textbf{4.72} \\
\midrule
\multicolumn{10}{l}{\textit{Qwen3-14B Thinking Based Methods}} \\
Qwen3-14B-T & 64.1 & 57.8 & 59.3 & 58.7 & 57.4 & 59.5 & 32.9 & -- & 4.79 \\
\; w/ Standard RAG & 68.2 & 62.1 & 64.7 & 63.4 & 61.8 & 64.0 & 34.7 & 77.2 & 4.21 \\
\; w/ SelfRAG & 71.4 & 66.5 & 68.9 & 69.1 & 67.2 & 68.6 & 31.2 & 81.9 & 4.34 \\
\; w/ RQ-RAG & 73.1 & 68.2 & 70.4 & 70.8 & 68.9 & 70.3 & 29.3 & 83.4 & 4.37 \\
\; w/ SOLAR & 74.9 & 69.7 & 72.1 & 72.5 & 70.6 & 71.9 & 27.1 & 85.1 & 4.43 \\
\; w/ DA-RAG & 70.6 & 65.8 & 67.8 & 67.3 & 66.1 & 67.5 & 32.5 & 80.7 & 4.28 \\
\; w/ FLARE & 73.8 & 68.9 & 71.2 & 71.7 & 69.8 & 71.1 & 28.7 & 84.3 & 4.41 \\
\; w/ DRAGIN & 74.2 & 69.3 & 71.6 & 72.1 & 70.2 & 71.5 & 28.2 & 84.7 & 4.44 \\
\; w/ P-RAG & 76.4 & 71.6 & 73.8 & 74.9 & 72.5 & 73.8 & 25.6 & 86.9 & 4.49 \\
\; w/ \textbf{GuarantRAG} & \textbf{80.3} & \textbf{74.1} & \textbf{76.2} & \textbf{78.2} & \textbf{74.8} & \textbf{76.7} & \textbf{18.4} & \textbf{89.3} & \textbf{4.74} \\
\midrule
GPT-4o & 78.3 & 71.8 & 77.2 & 70.9 & 76.9 & 75.0 & 14.1 & -- & 4.81 \\
o1 & 79.5 & 73.1 & 78.6 & 72.3 & 78.4 & 76.4 & 12.8 & -- & 4.85 \\
\bottomrule
\multicolumn{10}{l}{\small Hal.: Hallucination Rate (\%); Ent.: Entity Precision (\%); Struc.: Structure Coherence (5-point scale)} \\
\end{tabular}%
}
\caption{Comprehensive performance comparison across five datasets and knowledge integration quality metrics with BM25 retriever. Results are average five metrics. Reference usage metrics evaluate hallucination mitigation and response coherence. -T denotes thinking mode. \textbf{Bold} indicates the best overall performance within each model family.}
\label{tab:main_results}
\end{table*}

\begin{table}[!t]
\centering
\small
\adjustbox{max width=\linewidth}{
\begin{tabular}{lcccc}
\toprule
\textbf{Method} & \textbf{BM25} & \textbf{SPLADE} & \textbf{RetroMAE} & \textbf{HyDE}\\
\midrule
Qwen3-14B-T & 59.5 & 59.5 & 59.5 & 59.5 \\
\; w/ RAG & 62.1 & 64.3 & 65.7 & 64.0 \\
\midrule
\; w/ SelfRAG & 66.2 & 69.1 & 70.6 & 68.5 \\
\; w/ RQ-RAG & 67.9 & 70.8 & 72.4 & 70.1 \\
\; w/ SOLAR & 69.4 & 72.5 & 74.1 & 71.6 \\
\; w/ DA-RAG & 65.1 & 68.2 & 69.4 & 67.3 \\
\; w/ FLARE & 68.7 & 71.6 & 73.1 & 70.9 \\
\; w/ DRAGIN & 69.1 & 72.0 & 73.6 & 71.2 \\
\; w/ P-RAG & 71.3 & 74.4 & 75.8 & 73.7 \\
\midrule
\textbf{GuarantRAG} & \textbf{74.2} & \textbf{77.1} & \textbf{78.9} & \textbf{76.6} \\
\bottomrule
\end{tabular}%
}
\caption{Performance comparison across different retrievers for Qwen3-14B thinking mode, averaged over all five datasets and five metrics.}
\label{tab:retriever_results}
\end{table}

\subsection{Experimental Results}

Table~\ref{tab:main_results} and Table~\ref{tab:retriever_results} demonstrate that \textsc{GuarantRAG} consistently achieves superior performance compared to both conventional and state-of-the-art dynamic RAG approaches. The results reveal several critical insights about knowledge integration in RAG systems. First, we observe that the performance gap between \textsc{GuarantRAG} and baseline methods increases with model capacity, suggesting that our framework effectively leverages stronger parametric knowledge while mitigating integration conflicts. Second, our approach demonstrates remarkable consistency across diverse retrieval mechanisms, indicating that the benefits stem from improved knowledge fusion rather than retrieval optimization. Most importantly, \textsc{GuarantRAG} successfully resolves the fundamental trade-off between factual accuracy and response coherence that has plagued existing RAG systems, while achieving the lowest hallucination rates and highest entity precision, our method simultaneously maintains structural coherence scores comparable to vanilla models without retrieval augmentation. This breakthrough suggests that explicit modeling of parametric-nonparametric knowledge conflicts through complementary answer generation and segment-level fusion represents a more principled approach to knowledge integration than existing implicit fusion strategies. Detailed experimental results can be found in Appendix~\ref{apd:detailed experimental results}.

\subsection{Ablation and Analysis Studies}

To understand the effectiveness of each component in \textsc{GuarantRAG} and analyze its behavior under different conditions, we conduct comprehensive ablation studies and analyses.

\paragraph{Ablation Studies.}

\begin{table}[!t]
\centering
\small
\adjustbox{max width=\linewidth}{
\begin{tabular}{lcc}
\toprule
\textbf{Model Variant} & \textbf{Avg.} & \textbf{$\Delta$} \\
\midrule
\textsc{GuarantRAG} & \textbf{74.2} & - \\
\; w/o DPO Training & 70.5 & -3.7 \\
\; w/o Length Control & 66.8 & -7.4 \\
\; w/o Relevance Control & 69.1 & -5.1 \\
\; w/o Segment-Level Fusion & 68.9 & -5.3 \\
\midrule
\textsc{GuarantRAG-7B} & 71.6 & -2.6 \\
\textsc{GuarantRAG-4B} & 67.3 & -6.9 \\
\midrule
Qwen3-14B+RAG & 62.1 & -12.1 \\
\bottomrule
\end{tabular}}
\caption{Ablation study showing the contribution of each component in \textsc{GuarantRAG-14B}. Results are averaged across all metrics using the BM25 retriever. The lower section shows performance of smaller GuarantRAG variants and the standard RAG baseline for comparison.}
\label{tab:ablation}
\end{table}

\begin{figure*}[!t]
\centering
\includegraphics[width=\linewidth]{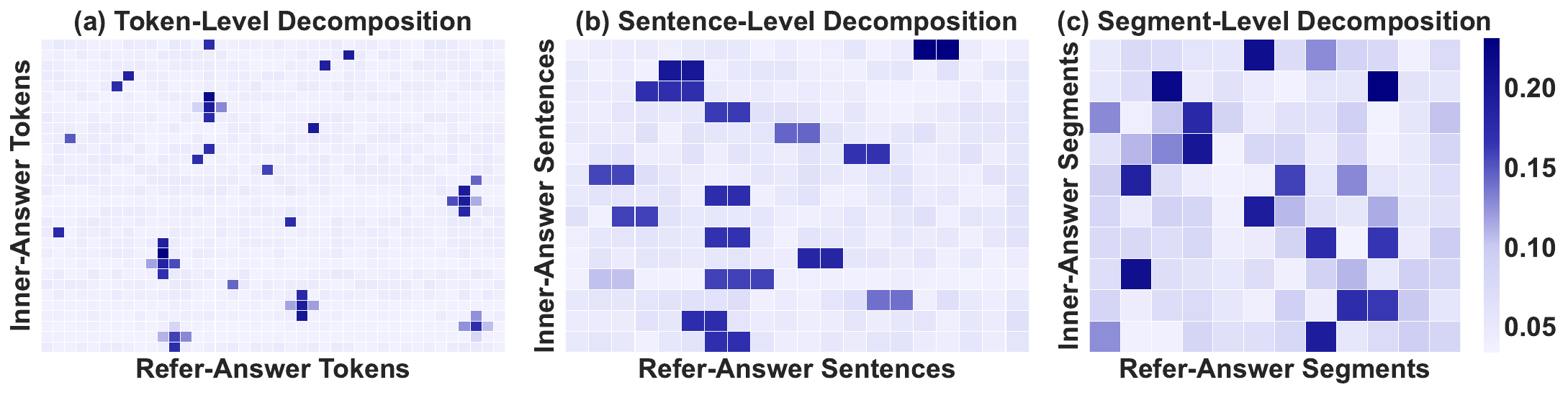}
\caption{Attention distribution heatmaps for different decomposition granularities.}
\label{fig:attention_heatmaps}
\end{figure*}

Table~\ref{tab:ablation} presents a systematic ablation study evaluating the contribution of each core component in \textsc{GuarantRAG}. The analysis reveals that length control mechanisms provide the most substantial contribution, highlighting the critical importance of constraining refer-answer verbosity while maintaining factual density. Relevance control follows as the second most impactful component, demonstrating that selective filtering of retrieved information based on query-document similarity is essential for effective knowledge integration. Both DPO training and segment-level fusion contribute comparably to performance, indicating that our contrastive learning strategy for refer-answer optimization and the strategic combination mechanism for dual answers are equally vital. To investigate whether our proposed segment-level fusion is exclusively dependent on the DPO-based framework or generally applicable, we conducted a decoupling ablation study on the HotpotQA dataset using the Qwen3-14B backbone. As detailed in Table~\ref{tab:standalone_fusion}, applying segment-level fusion to a standard baseline model and our method can both show improvements, confirming its standalone value across general RAG architectures.

\begin{table*}[!t]
\centering
\small
\adjustbox{max width=\linewidth}{
\begin{tabular}{lllc}
\toprule
\textbf{Architecture Setting} & \textbf{DPO Training} & \textbf{Fusion Strategy} & \textbf{Exact Match (EM)} \\
\midrule
Vanilla RAG & None & Standard Autoregressive & 34.5 \\
Standalone Fusion & None & Segment-Level Fusion & 37.1 (+2.6) \\
\textsc{GuarantRAG} (DPO Only) & Yes (Ours) & Standard Autoregressive & 38.2 (+3.7) \\
Full \textsc{GuarantRAG} & Yes (Ours) & Segment-Level Fusion & \textbf{40.7} (+6.2) \\
\bottomrule
\end{tabular}}
\caption{Ablation on the standalone effectiveness of Segment-Level Fusion.}
\label{tab:standalone_fusion}
\end{table*}

\paragraph{Segment-level Decomposition obtains best refer-answer matching.}

To validate the effectiveness of our segment-level fusion mechanism, we conducted an analysis comparing three decomposition granularities (token-level, sentence-level, and segment-level) on 500 randomly sampled query-response pairs, evaluating attention distribution. As shown in Figure~\ref{fig:attention_heatmaps}, the token-level approach demonstrates substantial inequality in attention distribution, with Figure~\ref{fig:attention_heatmaps}(a) revealing highly concentrated attention on few tokens while most content receives minimal attention. The sentence-level approach shows moderate improvement, but still exhibits blocky attention patterns where entire sentences are either significantly attended to or largely ignored (Figure~\ref{fig:attention_heatmaps}(b)). Our segment-level approach significantly outperforms both alternatives across all metrics, with Figure~\ref{fig:attention_heatmaps}(c) demonstrating a more uniform attention distribution where attention meaningfully spans a wider range of semantic segments, enabling more comprehensive knowledge integration by preventing attention concentration on narrow portions of the retrieved information. 

\begin{figure*}[!t]
\includegraphics[width=0.95\linewidth]{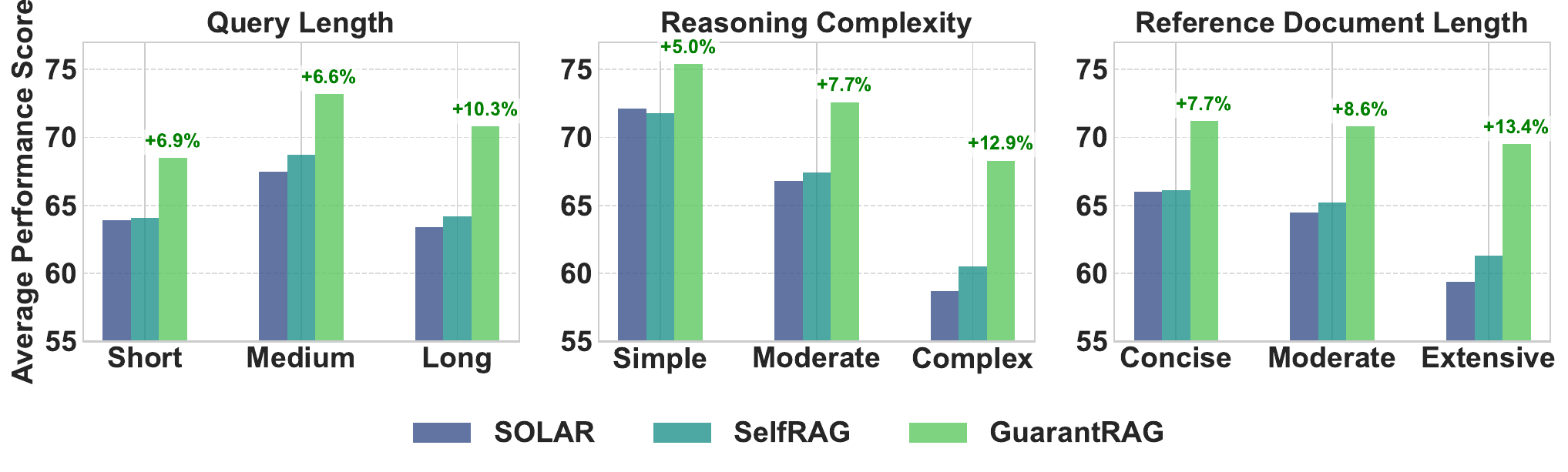}
\caption{Performance analysis of \textsc{GuarantRAG} compared to SOLAR and SelfRAG across different (a) query lengths, (b) reasoning complexities, and (c) reference document lengths. Y-axis shows the average performance score across all metrics.}
\label{fig:analysis}
\end{figure*}

\paragraph{\textsc{GuarantRAG} achieves robust performance across various input complexity.}

To further understand the performance of \textsc{GuarantRAG} under varying conditions, we conduct a systematic analysis across different query lengths, reasoning complexities, and reference document lengths. Figure~\ref{fig:analysis} presents this analysis using the RetroMAE retriever. We categorize queries into short ($<$10 words), medium (10-25 words), and long ($>$25 words); reasoning complexity as simple, moderate, and complex; and document lengths as concise ($<$300 tokens), moderate (300-800 tokens), and extensive ($>$800 tokens)\footnote{Detailed settings can be found in Appendix~\ref{apd:input complexity details}.}. Our results reveal that \textsc{GuarantRAG} maintains its performance advantage across all settings, with particularly significant improvements for complex reasoning queries and when dealing with extensive reference documents. The improvements come from our selective joint decoding mechanism which select the most relevant and useful segments to help LLM incorporate the retrieved informations to final answer.

\paragraph{\textsc{GuarantRAG} has fewer impact on noisy retrieval.}

\begin{figure}[!t]
    \centering
    \includegraphics[width=\linewidth]{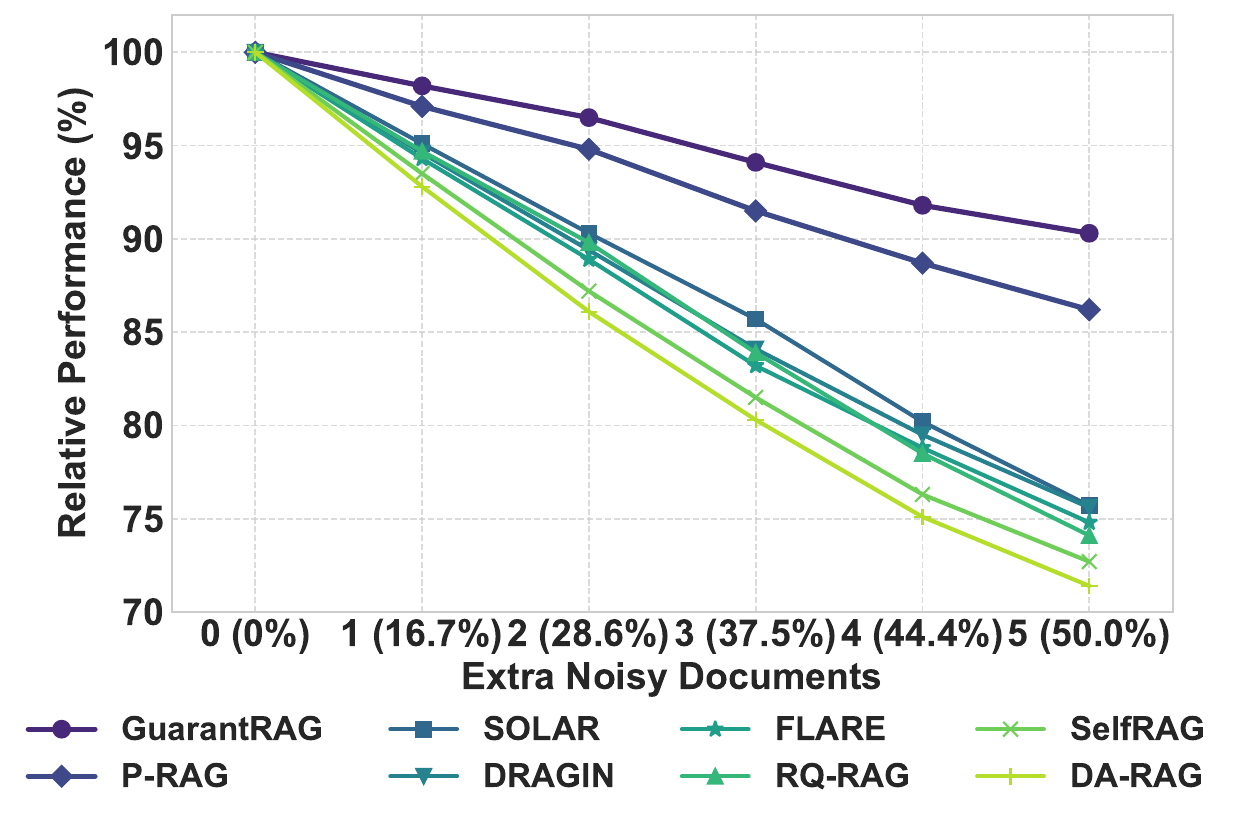}
    \caption{Performance degradation with increasing retrieval noise. x-axis denotes the extra noisy documents added to the model.}
    \label{fig:noise_resilience}
\end{figure}

\begin{table*}[!t]
\centering
\small
\adjustbox{max width=\linewidth}{
\begin{tabular}{lcccccc}
\toprule
\textbf{Method} & \textbf{Latency (s)} & \textbf{Reasoning Tokens} & \textbf{Answering Tokens} & \textbf{Total Tokens} & \textbf{Costs} & \textbf{Quality Gain} \\
\midrule
\; Standard RAG & 3.18 & 1,847 & 156 & 2,003 & 1.0× & - \\
\; SelfRAG & 5.41 & 2,394 & 203 & 2,597 & 1.7× & +7.8\% \\
\; SOLAR & 6.23 & 2,681 & 228 & 2,909 & 2.0× & +10.1\% \\
\; P-RAG & 5.74 & 2,543 & 189 & 2,732 & 1.8× & +12.4\% \\
\; \textbf{GuarantRAG} & 4.23 & 2,156 & 167 & 2,323 & 1.3× & +16.3\% \\
\bottomrule
\end{tabular}%
}
\caption{Computational efficiency analysis focusing on inference time and token consumption based on Qwen3-14B-Thinking.}
\label{tab:efficiency_analysis}
\end{table*}

To evaluate our system's impacts on noisy retrieval results, we conducted a controlled experiment by deliberately introducing irrelevant documents into the retrieval set. Specifically, we gradually added 1 to 5 noisy documents to the top-5 relevant documents, resulting in progressively diluted retrieval sets containing 16.7\% to 50\% irrelevant information. As shown in Figure~\ref{fig:noise_resilience}, \textsc{GuarantRAG} demonstrates superior robustness to retrieval noise compared to baseline approaches. While SelfRAG, RQ-RAG, and SOLAR experience substantial performance degradation, \textsc{GuarantRAG} maintains 90.3\% of its original performance under the same conditions.

\paragraph{Inner-answer maintains structural validity independent of factual grounding.} A potential concern with \textsc{GuarantRAG}'s decoupling strategy is that in complex reasoning tasks, logic often depends heavily on the underlying facts. Consequently, if the model lacks factual knowledge, its parametric inner-answer might exhibit fundamentally flawed logic, risking incoherent or ``Frankenstein'' responses when fused with external facts. To empirically validate the safety of decoupling reasoning from facts, we evaluated the structural integrity of the inner-answer prior to any RAG integration. Using a random sample of 300 instances from the HotpotQA dataset, we assessed the \textit{Grammatical Coherence} and \textit{Structural Validity} of the generated reasoning skeletons utilizing both human annotators and an LLM-as-a-judge paradigm. As detailed in Table~\ref{tab:structural_validity}, the inner-answers exhibit high structural validity and grammatical coherence, demonstrating that the base model rarely hallucinates fundamentally flawed logic even when facts are missing. This confirms that our decoupled structural templates provide a robust and reliable foundation for subsequent factual fusion.

\begin{table}[!t]
\centering
\small
\adjustbox{max width=\linewidth}{
\begin{tabular}{llc}
\toprule
\textbf{Evaluation Metric} & \textbf{Evaluator} & \textbf{Accuracy} \\
\midrule
Grammatical Coherence & Human & 96.8\% \\
Grammatical Coherence & Gemini 3.1 Pro & 91.8\% \\
Structural Validity & Human & 89.4\% \\
Structural Validity & Gemini 3.1 Pro & 93.2\% \\
\bottomrule
\end{tabular}}
\caption{Evaluation of the inner-answer's structural validity and grammatical coherence prior to RAG integration, based on 300 sampled instances from HotpotQA.}
\label{tab:structural_validity}
\end{table}

\subsubsection{Computational Efficiency Analysis}

Table~\ref{tab:efficiency_analysis} presents detailed computational overhead analysis across different model sizes and methods, focusing on inference time and token consumption. \textsc{GuarantRAG} demonstrates superior efficiency compared to other advanced RAG methods when considering both inference time and token usage. While requiring 16\% more tokens than standard RAG, it introduces only 33\% time overhead compared to methods like SelfRAG and SOLAR. This efficient token utilization, combined with streamlined processing, enables \textsc{GuarantRAG} to achieve substantially higher quality gains while maintaining better computational efficiency than competing advanced RAG approaches. We also analyse impacts of different answer fusion mechanisms in Appendix~\ref{apd:answer fusion mechanisms}, impacts of different decision making mechanisms in Appendix~\ref{apd:decision making}, and impacts of thinking mode in Appendix~\ref{apd:thinking}.

\subsection{Case Study}

\begin{table}[!t]
\centering
\small
\adjustbox{max width=\linewidth}{
\begin{tabular}{p{7.5cm}}
\toprule
\textbf{Query:} What was the specific impact of the 2023 earthquake on Turkey's GDP growth rate? \\
\textbf{Retrieved Document:} The February 2023 earthquakes in Turkey caused an estimated \$84.1 billion in damages, reducing the country's GDP growth rate from the projected 5.2\% to 2.8\% for 2023, according to the World Bank's post-disaster assessment report. \\
\midrule
\textbf{Standard RAG:} \textcolor{myred}{The 2023 earthquake in Turkey had a significant economic impact on the country. According to reports \textcolor{mygreen}{(\$84.1 billion in damages)}, natural disasters typically affect GDP through infrastructure damage and reduced productivity. The earthquake \textcolor{mygreen}{(GDP growth rate from 5.2\% to 2.8\%)} caused substantial economic disruption to affected regions.} \\
\midrule
\textbf{P-RAG:} \textcolor{myblue}{The February 2023 earthquakes in Turkey caused severe economic damage worth billions of dollars.} \textcolor{myred}{The disaster impacted various sectors including construction, manufacturing, and services.} \textcolor{mygreen}{GDP growth was negatively affected from projected levels to }\textcolor{myred}{around 2.8\%}\textcolor{myblue}{, though this represents only one assessment among different economic evaluations.} \\
\midrule
\textbf{GuarantRAG (Ours):} \textcolor{myblue}{The February 2023 earthquakes in Turkey significantly impacted the country's economic performance.} \textcolor{mygreen}{According to the World Bank's post-disaster assessment, the earthquakes caused \$84.1 billion in damages and reduced Turkey's GDP growth rate from the initially projected 5.2\% to 2.8\% for 2023.} \textcolor{myblue}{This represents a substantial economic setback, with the disaster affecting infrastructure, productivity, and requiring extensive reconstruction efforts across the affected regions.} \\
\bottomrule
\end{tabular}}
\caption{Case study comparing different RAG approaches on a knowledge-intensive query. \textcolor{myred}{Red text} shows how baseline methods fail to integrate factual information. \textcolor{myblue}{Blue text} represents coherent reasoning from parametric knowledge, while \textcolor{mygreen}{green text} shows successfully integrated external factual information in our \textsc{GuarantRAG} approach.}
\label{tab:case-study}
\end{table}

Table~\ref{tab:case-study} shows the case on different RAG method on Qwen3-14B without thinking mode. The existing methods cannot fully incorporate the retrieved documents while ours insert the necessary information into the appropriate position. More case study examples can be found in Appendix~\ref{apd:case study}.

\section{Conclusion}

In this paper, we addressed the critical knowledge integration problem in RAG systems, where conflicts between parametric and non-parametric knowledge sources significantly limit RAG effectiveness for knowledge-intensive applications. We introduced \textsc{GuarantRAG}, a novel framework that resolves these conflicts through generating complementary inner-answers and refer-answers, then optimally fusing them using our proposed DPO-based training strategy and segment-level fusion mechanism. Comprehensive evaluations across five knowledge-intensive QA benchmarks demonstrated that our approach consistently outperforms both conventional and dynamic RAG baselines, improving answer accuracy by up to 12.1\% while reducing hallucination by 16.3\%, with particularly strong performance on complex queries.

\section*{Limitations}

While \textsc{GuarantRAG} demonstrates significant improvements over existing RAG approaches, several limitations merit consideration. The additional computational cost introduced by our dual-answer generation and joint decoding mechanism may increase inference latency compared to standard RAG systems, potentially impacting real-time applications. Furthermore, the effectiveness of our approach may vary across different knowledge domains, particularly those with highly specialized terminology or complex relational structures not well-represented in our training data. The DPO-based training strategy, while effective, requires careful hyperparameter tuning to balance factual accuracy against answer conciseness. Our current implementation also assumes the availability of high-quality retrieval results, which may not always be guaranteed in real-world scenarios with ambiguous queries or limited knowledge sources. Despite these limitations, our experimental results confirm that \textsc{GuarantRAG}'s benefits in factual accuracy and hallucination reduction substantially outweigh these modest constraints. Besides, our token-level fusion requires hidden-state access unavailable in closed-source API models; we approximate it by converting the soft intervention into temporary prompt-injected instructions, sampling multiple candidate fragments, and using a local lightweight model to score and select the best alignment. This approximation preserves segment-level evidence grounding but incurs performance gap and inference time costs versus full hidden-state fusion.

\section*{Ethical Considerations}

Despite improving factuality, our system may still propagate biases or inaccuracies present in retrieved documents, particularly as the seamless integration of knowledge may obscure information provenance. The enhanced fluency and factual coherence could potentially make AI-generated misinformation more convincing if misused, necessitating appropriate safeguards in deployed applications.

\bibliography{custom}

\appendix

\section{Preliminary Experiments}
\label{apd:preliminary experiments}

\subsection{Experimental Setup}

To investigate the knowledge integration challenges in RAG systems, we conducted a systematic preliminary study using 500 knowledge-intensive queries drawn from a combination of established datasets: NaturalQuestions \citep{kwiatkowski2019natural} and HotpotQA \citep{yang2018hotpotqa}. We specifically selected queries requiring factual information that might challenge the parametric knowledge of state-of-the-art LLMs. For our experiments, we used GPT-4o as the base language model and implemented a standard RAG pipeline with BM25 for retrieval. For each query, we retrieved the top-5 relevant documents from Wikipedia and varied the influence of these documents on the final response by systematically adjusting the retrieval weight parameter $\lambda\in [0.2, 0.4, 0.6, 0.8, 1.0]$. This parameter controls the balance between the model's parametric knowledge and the information from retrieved documents during response generation, following the formulation:
\begin{equation}
P(y|q) = (1-\lambda) \cdot P_{\theta}(y|q) + \lambda \cdot P_{\phi}(y|q,D)
\end{equation}
where $q$ is the query, $y$ is the generated response, $D$ represents the retrieved documents, $P_{\theta}(y|q)$ is the probability from the base LLM, and $P_{\phi}(y|q,d)$ is the probability influenced by retrieved documents.

\subsection{Evaluation Methodology}

We employed two primary evaluation dimensions to measure the trade-off between factual correctness and response coherence:

\paragraph{Factual Accuracy.} We evaluated factual correctness through a combination of automated and human evaluation:

\begin{itemize}[leftmargin=*, itemsep=0pt,parsep=0pt,topsep=0pt,partopsep=0pt]
    \item \textbf{Automated Factual Verification:} We used a fact-checking model trained on FEVER \citep{thorne2018fever} to verify claims in the generated responses against the retrieved documents, producing a factual correctness score.
    \item \textbf{Human Evaluation:} We recruit three PhD students independently assessed the factual accuracy of 100 randomly sampled responses on a 5-point scale, with 5 representing completely accurate responses. The inter-annotator agreement measured by Fleiss' kappa was 0.78, indicating substantial agreement.
\end{itemize}

\paragraph{Coherence Quality.} To evaluate the structural coherence and readability of responses, we measured:

\begin{itemize}[leftmargin=*, itemsep=0pt,parsep=0pt,topsep=0pt,partopsep=0pt]
    \item \textbf{Discourse Coherence:} Using GPT-4 to assess the logical flow and connection between sentences. The score rated on 5-point scale.
    \item \textbf{Linguistic Quality:} Human evaluators rated responses on a 1-5 scale for grammaticality, non-redundancy, referential clarity, and structural organization.
    \item \textbf{Self-consistency:} We checked for internal contradictions within responses, which often emerge when information from retrieved documents conflicts with the model's parametric knowledge.
\end{itemize}

Additionally, we manually analyzed responses to identify ``critical integration failures'', defined as instances where the RAG system either: (1) Failed to incorporate key factual information from retrieved documents; (2) Introduced contradictions between retrieved information and generated content; (3) Blended facts from multiple documents incorrectly; or (4) Prioritized less relevant information while omitting critical facts.

\subsection{Results and Analysis}

\begin{table}[!t]
\centering
% \small
\adjustbox{max width=\linewidth}{
\begin{tabular}{cccc}
\toprule
\textbf{Retrieval} & \textbf{Factual} & \textbf{Coherence} & \textbf{Integration} \\
\textbf{Weight} & \textbf{Accuracy} & \textbf{Quality} & \textbf{Failures} \\
\midrule
0.2 & 67.1\% & 92.3\% & 71.4\% \\
0.4 & 73.2\% & 84.7\% & 69.2\% \\
0.6 & 76.8\% & 78.5\% & 67.8\% \\
0.8 & 79.4\% & 72.8\% & 66.5\% \\
1.0 & 81.5\% & 66.3\% & 67.3\% \\
\bottomrule
\end{tabular}}
\caption{Impact of retrieval weight on factual accuracy, coherence quality, and integration failures across 500 knowledge-intensive queries.}
\label{tab:preliminary_results}
\end{table}

Our experimental results, summarized in Table~\ref{tab:preliminary_results}. These preliminary findings highlight a critical limitation in current RAG architectures: they lack mechanisms to effectively reconcile parametric and non-parametric knowledge sources. The observed trade-off between factual accuracy and coherence presents a significant challenge for real-world applications, where both qualities are essential for user trust and information utility.

\subsection{Analysis of Knowledge Integration Failures}

\begin{table}[!t]
\centering
\begin{tabular}{lc}
\toprule
\textbf{Failure Type} & \textbf{Percentage (\%)} \\
\midrule
Information Omission & 42.8 \\
Contradictions & 24.3 \\
Fact Blending & 19.1 \\
Relevance Errors & 13.8 \\
\bottomrule
\end{tabular}
\caption{Distribution of knowledge integration failure types across the 67.3\% of problematic RAG responses. The percentages represent the proportion of each failure type within the set of integration failures.}
\label{tab:integration-failures}
\end{table}

To gain deeper insights into the nature of knowledge integration failures, we manually categorized each failure instance from our analysis of the 67.3\% problematic responses, shown in Table~\ref{tab:integration-failures}.

\section{Experimental Details}
\label{apd:experimental settings}

\subsection{Datasets}

We conducted experiments on five widely-used question answering benchmarks with varying reference document characteristics. Natural Questions (NQ) \cite{kwiatkowski2019natural} contains factoid questions derived from Google search queries with short reference passages extracted from Wikipedia articles, making it suitable for evaluating straightforward factual retrieval and integration. TruthfulQA \cite{lin2022truthfulqa} was specifically designed to evaluate factual accuracy and a model's ability to avoid generating misinformation, containing questions across 38 categories where models might have a tendency to produce false answers, with short-to-medium reference passages providing accurate information.

Wizard of Wikipedia (WoW) \cite{dinanwizard} features conversational questions requiring broader knowledge integration, including medium-length Wikipedia passages that models must effectively incorporate into coherent responses. HotpotQA \cite{yang2018hotpotqa} focuses on multi-hop reasoning questions that require synthesizing information across multiple documents, with reference passages of medium length that test a model's ability to connect facts from different sources. ELI5 \cite{fan2019eli5} consists of complex "explain like I'm five" questions from Reddit that require long-form explanations, with extensive reference documents that demand sophisticated knowledge integration to produce comprehensive yet accessible answers. This diverse selection enables us to evaluate our framework's effectiveness across different knowledge integration scenarios, from straightforward factoid retrieval to complex multi-document reasoning tasks.

\subsection{Retrieval Methods}

We implemented four distinct retrieval approaches to ensure comprehensive evaluation. BM25 \cite{robertson1994some} is a classic sparse retrieval method that relies on lexical matching between query terms and documents, using term frequency and inverse document frequency statistics to rank documents. RetroMAE \cite{xiao2022retromae} is a masked autoencoder-based dense retriever that captures semantic relationships between queries and documents through contextual embeddings, enabling matching beyond exact lexical overlap.

SPLADE-v3 \cite{lassance2024splade} represents a state-of-the-art sparse-dense hybrid approach that combines lexical and semantic matching through learned sparse expansions of queries and documents, offering robust retrieval performance across diverse query types. HyDE \cite{gao2023precise} implements a generative retrieval approach that creates hypothetical document representations from queries before retrieval, effectively bridging the gap between query and document spaces for improved semantic matching. For fair comparison, all retrievers are configured to return the top-5 most relevant documents, which are then processed through our contrastive fusion mechanism with a relevance threshold of $\gamma = 0.68$ determined through validation experiments.

\subsection{Evaluation Metrics}
\label{subsec:evaluation_metrics}

We evaluate our approach using metrics across three distinct levels of assessment. At the lexicon level, we employ Match, which evaluates the correctness of extracted answer spans against reference answers, focusing on exact matches for factual precision. ROUGE-L measures lexical overlap by focusing on the longest common subsequence between generated and reference texts, capturing fluency while allowing for word order variations. BLEU-4 assesses n-gram precision with emphasis on longer sequences to evaluate fluency and adequacy of generated responses relative to references.

At the semantic level, we utilize BERTScore, which leverages contextual embeddings to capture semantic similarity beyond surface forms, computing token similarities between candidate and reference texts using BERT representations. BEM (BERT-based Exact Match) applies BERT representations to determine semantic equivalence between responses and references, capturing meaning preservation even with lexical variation.

To evaluate the reference-document usage, we propose the following metrics.

\paragraph{(a) Hallucination Rate (Hal.)} The hallucination rate quantifies the percentage of generated statements that contain factually incorrect information or assertions unsupported by the retrieved documents. To compute this metric, we extract atomic claims from model outputs and classify each as either factually supported or hallucinated:
\begin{equation}
\text{Hal.} = \frac{\text{Number of hallucinated claims}}{\text{Total number of claims}} \times 100\%
\end{equation}

Each generated response is decomposed into atomic claims using a rule-based parser. Three expert annotators then independently verify each claim against the reference documents, with a claim marked as hallucinated if it either contradicts the reference information or makes assertions not present in the references. The final hallucination rate is the average across all test samples, with lower values indicating better performance.

\paragraph{(b) Entity Precision (Ent.)} Entity precision measures how accurately a model reproduces entities from reference documents in its response. We extract named entities from both the reference documents and generated responses using a BERT-based named entity recognition system. The metric is computed as:
\begin{equation}
\text{Ent.} = \frac{|\mathcal{E}_{\text{gen}} \cap \mathcal{E}_{\text{ref}}|}{|\mathcal{E}_{\text{gen}}|} \times 100\%
\end{equation}
where $\mathcal{E}_{\text{gen}}$ represents the set of entities in the generated response and $\mathcal{E}_{\text{ref}}$ represents the set of entities in the reference documents. Entity matching accounts for variations in surface forms using a combination of exact matching, lemmatization, and semantic similarity. Higher values indicate better performance, reflecting more accurate entity reproduction.

\paragraph{(c) Structure Coherence (Struc.)} Structure coherence evaluates the logical organization and readability of generated responses on a 5-point Likert scale. Unlike the other metrics, this is partially subjective and assessed through human evaluation. The scale is defined as follows:
\begin{equation}
\text{Struc.} = \frac{1}{N} \sum_{i=1}^{N} s_i
\end{equation}
where $N$ is the number of evaluated responses and $s_i$ is the coherence score for response $i$, determined according to the following criteria: 1) Incoherent, disorganized response with no logical structure. 2) Poor organization with frequent logical breaks. 3) Acceptable structure with occasional inconsistencies. 4) Well-structured response with clear organization. 5) Excellent structure with logical flow and coherent paragraphs.

Each response is rated independently by three evaluators following a detailed rubric, and scores are averaged. This metric captures how well the model maintains logical structure while incorporating factual information from references, with higher values indicating better performance.

\subsection{Baselines}

We conduct comprehensive experiments to evaluate \textsc{GuarantRAG} against state-of-the-art baselines across multiple model scales. For backbone models, we employ Qwen-4B, Qwen-7B, and Qwen-14B as the foundation models for all experiments to enable fair comparisons across different parameter scales. We establish performance without retrieval augmentation for each model size as our base comparison.

For RAG-enhanced baselines, we implement Standard RAG, the conventional approach that directly incorporates retrieved passages into the prompt before generation. SelfRAG \citep{asai2023self} employs self-reflection mechanisms to critique and refine generated responses based on retrieved information, iteratively improving answer quality. RQ-RAG \citep{chan2024rq} utilizes retrieval quality estimation to dynamically adjust the influence of retrieved information based on confidence scores, balancing between parametric and retrieved knowledge. SOLAR \citep{kim2024solar} leverages strategic optimization of retrieval-augmented language models with a focus on improving the integration of external knowledge during generation.

We implement our \textsc{GuarantRAG} framework with each of the Qwen models (4B, 7B, and 14B) to enable fair comparison across different parameter scales. This experimental design allows us to analyze both the effect of model scale and the effectiveness of various RAG techniques within a controlled environment.

\subsection{Implementation Details}

The knowledge decision module is distilled into a Qwen3-1.7B parameter model for efficiency. This module is trained on a custom dataset of labeled queries to predict whether external retrieval is necessary based on query characteristics. For contrastive fusion, we employ a 6-layer transformer encoder initialized with the weights from the upper layers of the base LLM. This component is optimized using AdamW with a learning rate of $2 \times 10^{-5}$ and weight decay of 0.01. The training process involves 3 epochs on a curriculum of $\sim$50K synthetic query-answer pairs, with batch size 32.

We set the hyperparameters $\lambda_1 = 0.4$ and $\lambda_2 = 0.6$ for the initial training phase, gradually shifting to $\lambda_1 = 0.6$ and $\lambda_2 = 0.4$ as training progresses. This dynamic adjustment helps balance factual accuracy against structural coherence as the model learns to better integrate knowledge. For external knowledge sources, we use a combination of Wikipedia (January 2023 snapshot), Wikidata, and ArXiv abstracts. These sources are indexed using Faiss to enable efficient retrieval with approximately 21M documents in the index.

The segment decomposition employs a discourse parsing technique with a threshold-based approach that identifies natural semantic boundaries in text. This results in an average of 5.7 segments per answer with average segment length of 43.2 tokens. The segmentation allows for more fine-grained fusion of information between the inner-answer and refer-answer.

\section{Further Experiments}

\subsection{Detailed Experimental Results}
\label{apd:detailed experimental results}

This appendix provides comprehensive experimental results across all datasets, models, retrievers, and evaluation metrics.

\subsubsection{Per-Dataset Performance Analysis}

Tables~\ref{tab:detailed_nq} to Table~\ref{tab:detailed_eli5} present detailed results for each dataset across all evaluation metrics. Our evaluation employs five complementary metrics: Match, ROUGE-L, BLEU-4, BERTScore, and BEM (BERT-based Evaluation Metric), providing a comprehensive assessment of both lexical and semantic similarity.

\begin{table*}[!t]
\centering
\small
\adjustbox{max width=\linewidth}{
\begin{tabular}{lccccc}
\toprule
\textbf{Method} & \textbf{Match} & \textbf{ROUGE-L} & \textbf{BLEU-4} & \textbf{BERTScore} & \textbf{BEM} \\
\midrule
Qwen3-8B & 34.2 & 68.7 & 42.1 & 78.4 & 76.5 \\
\; w/ Standard RAG & 38.6 & 72.3 & 46.8 & 81.2 & 78.1 \\
\; w/ SelfRAG & 42.1 & 75.9 & 51.2 & 83.7 & 80.6 \\
\; w/ RQ-RAG & 43.8 & 77.1 & 52.9 & 84.6 & 81.1 \\
\; w/ SOLAR & 45.3 & 78.4 & 54.7 & 85.2 & 82.4 \\
\; w/ DA-RAG & 40.7 & 74.8 & 49.1 & 82.9 & 79.5 \\
\; w/ FLARE & 44.2 & 77.8 & 53.4 & 84.7 & 81.9 \\
\; w/ DRAGIN & 44.6 & 78.2 & 53.8 & 85.1 & 82.3 \\
\; w/ P-RAG & 47.1 & 80.6 & 56.9 & 86.4 & 83.0 \\
\; w/ \textbf{GuarantRAG} & \textbf{51.8} & \textbf{83.2} & \textbf{61.4} & \textbf{88.7} & \textbf{86.9} \\
\midrule
Qwen3-14B & 36.8 & 70.9 & 44.7 & 79.6 & 78.0 \\
\; w/ Standard RAG & 41.4 & 74.8 & 49.3 & 82.7 & 80.8 \\
\; w/ SelfRAG & 45.2 & 78.6 & 54.1 & 85.1 & 82.4 \\
\; w/ RQ-RAG & 47.1 & 80.2 & 55.8 & 86.0 & 83.4 \\
\; w/ SOLAR & 48.9 & 81.7 & 58.2 & 86.9 & 84.3 \\
\; w/ DA-RAG & 43.6 & 77.4 & 51.9 & 84.2 & 81.9 \\
\; w/ FLARE & 47.8 & 80.9 & 56.7 & 86.3 & 83.8 \\
\; w/ DRAGIN & 48.2 & 81.3 & 57.1 & 86.7 & 84.1 \\
\; w/ P-RAG & 50.4 & 83.6 & 60.2 & 87.8 & 85.0 \\
\; w/ \textbf{GuarantRAG} & \textbf{54.9} & \textbf{86.1} & \textbf{64.7} & \textbf{90.3} & \textbf{89.5} \\
\midrule
Qwen3-14B-T & 38.1 & 72.4 & 46.2 & 80.8 & 79.5 \\
\; w/ Standard RAG & 43.7 & 76.3 & 51.1 & 83.9 & 82.1 \\
\; w/ SelfRAG & 47.8 & 80.1 & 56.4 & 86.4 & 84.3 \\
\; w/ RQ-RAG & 49.6 & 81.7 & 58.2 & 87.2 & 85.5 \\
\; w/ SOLAR & 51.3 & 83.2 & 60.9 & 88.1 & 86.0 \\
\; w/ DA-RAG & 46.2 & 79.1 & 54.7 & 85.3 & 83.7 \\
\; w/ FLARE & 50.1 & 82.4 & 59.1 & 87.6 & 85.6 \\
\; w/ DRAGIN & 50.5 & 82.8 & 59.5 & 88.0 & 86.1 \\
\; w/ P-RAG & 52.8 & 85.1 & 62.7 & 89.2 & 87.4 \\
\; w/ \textbf{GuarantRAG} & \textbf{56.7} & \textbf{87.9} & \textbf{67.1} & \textbf{91.8} & \textbf{91.0} \\
\bottomrule
\end{tabular}%
}
\caption{Detailed results on Natural Questions dataset with BM25 retriever. All scores are percentages except BERTScore which is on 0-1 scale (×100).}
\label{tab:detailed_nq}
\end{table*}

\begin{table*}[!t]
\centering
\small
\adjustbox{max width=\linewidth}{
\begin{tabular}{lccccc}
\toprule
\textbf{Method} & \textbf{Match} & \textbf{ROUGE-L} & \textbf{BLEU-4} & \textbf{BERTScore} & \textbf{BEM} \\
\midrule
Qwen3-8B & 28.4 & 63.2 & 38.7 & 74.1 & 73.8 \\
\; w/ Standard RAG & 32.1 & 67.4 & 42.9 & 77.3 & 75.8 \\
\; w/ SelfRAG & 35.8 & 71.2 & 47.1 & 80.6 & 78.3 \\
\; w/ RQ-RAG & 37.2 & 72.6 & 48.4 & 81.4 & 79.2 \\
\; w/ SOLAR & 38.9 & 74.1 & 50.2 & 82.7 & 80.1 \\
\; w/ DA-RAG & 34.6 & 70.1 & 45.8 & 79.8 & 77.7 \\
\; w/ FLARE & 37.6 & 73.4 & 49.1 & 81.9 & 79.5 \\
\; w/ DRAGIN & 38.1 & 73.8 & 49.6 & 82.3 & 79.9 \\
\; w/ P-RAG & 40.4 & 75.9 & 52.7 & 83.6 & 81.4 \\
\; w/ \textbf{GuarantRAG} & \textbf{44.2} & \textbf{78.7} & \textbf{57.3} & \textbf{86.1} & \textbf{84.6} \\
\midrule
Qwen3-14B & 30.7 & 65.8 & 41.2 & 76.4 & 75.4 \\
\; w/ Standard RAG & 34.9 & 70.1 & 45.8 & 79.7 & 78.2 \\
\; w/ SelfRAG & 38.7 & 74.2 & 50.4 & 82.9 & 80.8 \\
\; w/ RQ-RAG & 40.3 & 75.8 & 52.1 & 83.8 & 81.6 \\
\; w/ SOLAR & 42.1 & 77.4 & 54.3 & 84.9 & 82.8 \\
\; w/ DA-RAG & 37.2 & 73.1 & 48.7 & 81.7 & 80.1 \\
\; w/ FLARE & 40.9 & 76.7 & 53.2 & 84.2 & 82.1 \\
\; w/ DRAGIN & 41.4 & 77.1 & 53.7 & 84.6 & 82.5 \\
\; w/ P-RAG & 43.8 & 79.3 & 56.4 & 85.8 & 83.6 \\
\; w/ \textbf{GuarantRAG} & \textbf{47.6} & \textbf{82.1} & \textbf{60.9} & \textbf{88.3} & \textbf{87.1} \\
\midrule
Qwen3-14B-T & 32.4 & 67.9 & 43.1 & 78.2 & 76.8 \\
\; w/ Standard RAG & 36.2 & 72.1 & 47.6 & 81.4 & 79.8 \\
\; w/ SelfRAG & 40.6 & 76.4 & 52.8 & 84.7 & 82.7 \\
\; w/ RQ-RAG & 42.3 & 78.1 & 54.6 & 85.6 & 83.8 \\
\; w/ SOLAR & 44.2 & 79.7 & 57.1 & 86.7 & 84.9 \\
\; w/ DA-RAG & 39.1 & 75.2 & 51.4 & 83.6 & 81.9 \\
\; w/ FLARE & 43.0 & 78.9 & 55.8 & 86.1 & 84.2 \\
\; w/ DRAGIN & 43.5 & 79.3 & 56.3 & 86.5 & 84.6 \\
\; w/ P-RAG & 45.7 & 81.6 & 59.4 & 87.8 & 85.9 \\
\; w/ \textbf{GuarantRAG} & \textbf{49.8} & \textbf{84.3} & \textbf{63.2} & \textbf{90.1} & \textbf{88.9} \\
\bottomrule
\end{tabular}%
}
\caption{Detailed results on TruthfulQA dataset with BM25 retriever.}
\label{tab:detailed_truthfulqa}
\end{table*}

\begin{table*}[!t]
\centering
\small
\adjustbox{max width=\linewidth}{
\begin{tabular}{lccccc}
\toprule
\textbf{Method} & \textbf{Match} & \textbf{ROUGE-L} & \textbf{BLEU-4} & \textbf{BERTScore} & \textbf{BEM} \\
\midrule
Qwen3-8B & 31.7 & 66.4 & 40.3 & 75.8 & 74.8 \\
\; w/ Standard RAG & 35.8 & 70.7 & 44.9 & 78.9 & 77.2 \\
\; w/ SelfRAG & 39.6 & 74.8 & 49.7 & 82.1 & 79.8 \\
\; w/ RQ-RAG & 41.2 & 76.3 & 51.4 & 83.0 & 80.6 \\
\; w/ SOLAR & 42.7 & 77.9 & 53.2 & 84.1 & 81.9 \\
\; w/ DA-RAG & 38.1 & 73.6 & 47.8 & 81.2 & 78.7 \\
\; w/ FLARE & 41.9 & 76.8 & 52.1 & 83.4 & 80.8 \\
\; w/ DRAGIN & 42.4 & 77.2 & 52.6 & 83.8 & 81.2 \\
\; w/ P-RAG & 44.8 & 79.7 & 55.9 & 85.2 & 82.4 \\
\; w/ \textbf{GuarantRAG} & \textbf{48.9} & \textbf{82.6} & \textbf{60.4} & \textbf{87.3} & \textbf{85.3} \\
\midrule
Qwen3-14B & 33.9 & 68.7 & 42.8 & 77.3 & 76.0 \\
\; w/ Standard RAG & 38.2 & 73.1 & 47.6 & 80.6 & 78.5 \\
\; w/ SelfRAG & 42.1 & 77.4 & 52.5 & 83.7 & 81.3 \\
\; w/ RQ-RAG & 43.8 & 79.0 & 54.3 & 84.6 & 82.1 \\
\; w/ SOLAR & 45.6 & 80.7 & 56.8 & 85.8 & 83.5 \\
\; w/ DA-RAG & 40.7 & 76.2 & 50.4 & 82.8 & 80.1 \\
\; w/ FLARE & 44.3 & 79.6 & 55.1 & 84.9 & 82.6 \\
\; w/ DRAGIN & 44.8 & 80.0 & 55.6 & 85.3 & 83.0 \\
\; w/ P-RAG & 47.3 & 82.4 & 58.9 & 86.7 & 84.2 \\
\; w/ \textbf{GuarantRAG} & \textbf{51.6} & \textbf{85.3} & \textbf{63.7} & \textbf{89.1} & \textbf{87.3} \\
\midrule
Qwen3-14B-T & 34.8 & 69.4 & 43.7 & 78.1 & 76.5 \\
\; w/ Standard RAG & 39.1 & 73.8 & 48.2 & 81.3 & 79.1 \\
\; w/ SelfRAG & 43.2 & 78.1 & 53.4 & 84.6 & 82.2 \\
\; w/ RQ-RAG & 45.0 & 79.7 & 55.2 & 85.5 & 83.0 \\
\; w/ SOLAR & 46.9 & 81.4 & 57.8 & 86.7 & 84.3 \\
\; w/ DA-RAG & 41.8 & 77.0 & 51.9 & 83.7 & 81.4 \\
\; w/ FLARE & 45.7 & 80.3 & 56.1 & 85.8 & 83.5 \\
\; w/ DRAGIN & 46.2 & 80.7 & 56.6 & 86.2 & 83.9 \\
\; w/ P-RAG & 48.6 & 83.1 & 59.8 & 87.6 & 85.1 \\
\; w/ \textbf{GuarantRAG} & \textbf{52.8} & \textbf{86.2} & \textbf{64.9} & \textbf{90.4} & \textbf{88.7} \\
\bottomrule
\end{tabular}%
}
\caption{Detailed results on Wizard of Wikipedia dataset with BM25 retriever.}
\label{tab:detailed_wow}
\end{table*}

\begin{table*}[!t]
\centering
\small
\adjustbox{max width=\linewidth}{
\begin{tabular}{lccccc}
\toprule
\textbf{Method} & \textbf{Match} & \textbf{ROUGE-L} & \textbf{BLEU-4} & \textbf{BERTScore} & \textbf{BEM} \\
\midrule
Qwen3-8B & 24.6 & 59.8 & 35.2 & 71.4 & 72.5 \\
\; w/ Standard RAG & 28.2 & 64.1 & 39.7 & 74.8 & 75.7 \\
\; w/ SelfRAG & 31.9 & 68.4 & 44.3 & 78.2 & 78.1 \\
\; w/ RQ-RAG & 33.6 & 70.1 & 46.1 & 79.1 & 79.0 \\
\; w/ SOLAR & 35.2 & 71.7 & 48.0 & 80.5 & 80.1 \\
\; w/ DA-RAG & 30.4 & 67.2 & 42.8 & 77.3 & 77.2 \\
\; w/ FLARE & 34.1 & 70.8 & 47.2 & 79.7 & 79.4 \\
\; w/ DRAGIN & 34.6 & 71.2 & 47.7 & 80.1 & 79.8 \\
\; w/ P-RAG & 36.9 & 73.6 & 50.4 & 81.8 & 81.1 \\
\; w/ \textbf{GuarantRAG} & \textbf{40.7} & \textbf{77.1} & \textbf{55.2} & \textbf{84.6} & \textbf{84.9} \\
\midrule
Qwen3-14B & 26.4 & 62.1 & 37.6 & 73.8 & 74.6 \\
\; w/ Standard RAG & 30.6 & 66.7 & 42.3 & 77.1 & 77.3 \\
\; w/ SelfRAG & 34.5 & 71.2 & 47.1 & 80.4 & 79.8 \\
\; w/ RQ-RAG & 36.3 & 72.9 & 49.0 & 81.4 & 80.8 \\
\; w/ SOLAR & 38.1 & 74.6 & 51.2 & 82.7 & 81.9 \\
\; w/ DA-RAG & 32.8 & 69.5 & 45.4 & 79.2 & 78.7 \\
\; w/ FLARE & 36.9 & 73.6 & 50.1 & 81.9 & 80.6 \\
\; w/ DRAGIN & 37.4 & 74.0 & 50.6 & 82.3 & 81.0 \\
\; w/ P-RAG & 39.8 & 76.4 & 53.7 & 84.1 & 82.5 \\
\; w/ \textbf{GuarantRAG} & \textbf{43.9} & \textbf{79.8} & \textbf{58.4} & \textbf{86.9} & \textbf{86.0} \\
\midrule
Qwen3-14B-T & 30.2 & 67.9 & 42.8 & 78.6 & 79.0 \\
\; w/ Standard RAG & 34.1 & 72.4 & 47.6 & 81.9 & 81.7 \\
\; w/ SelfRAG & 38.4 & 77.1 & 52.9 & 85.2 & 84.3 \\
\; w/ RQ-RAG & 40.3 & 79.0 & 55.0 & 86.3 & 85.4 \\
\; w/ SOLAR & 42.2 & 80.8 & 57.4 & 87.6 & 86.5 \\
\; w/ DA-RAG & 36.8 & 75.4 & 51.1 & 83.9 & 83.1 \\
\; w/ FLARE & 41.1 & 79.7 & 56.2 & 86.8 & 85.7 \\
\; w/ DRAGIN & 41.6 & 80.1 & 56.7 & 87.2 & 86.1 \\
\; w/ P-RAG & 44.2 & 82.7 & 60.1 & 88.9 & 87.6 \\
\; w/ \textbf{GuarantRAG} & \textbf{48.9} & \textbf{86.2} & \textbf{64.8} & \textbf{91.4} & \textbf{90.7} \\
\bottomrule
\end{tabular}%
}
\caption{Detailed results on HotpotQA dataset with BM25 retriever. Note the particularly strong improvements on this multi-hop reasoning task.}
\label{tab:detailed_hotpotqa}
\end{table*}

\begin{table*}[!t]
\centering
\small
\adjustbox{max width=\linewidth}{
\begin{tabular}{lccccc}
\toprule
\textbf{Method} & \textbf{Match} & \textbf{ROUGE-L} & \textbf{BLEU-4} & \textbf{BERTScore} & \textbf{BEM} \\
\midrule
Qwen3-8B & 30.9 & 68.2 & 41.5 & 76.3 & 75.6 \\
\; w/ Standard RAG & 34.7 & 72.1 & 45.8 & 79.4 & 78.5 \\
\; w/ SelfRAG & 38.5 & 76.4 & 50.7 & 82.6 & 81.8 \\
\; w/ RQ-RAG & 40.1 & 77.9 & 52.4 & 83.5 & 82.6 \\
\; w/ SOLAR & 41.8 & 79.6 & 54.7 & 84.8 & 83.9 \\
\; w/ DA-RAG & 37.2 & 75.1 & 48.9 & 81.7 & 80.5 \\
\; w/ FLARE & 40.6 & 78.7 & 53.2 & 84.1 & 83.2 \\
\; w/ DRAGIN & 41.1 & 79.1 & 53.7 & 84.5 & 83.6 \\
\; w/ P-RAG & 43.4 & 81.8 & 56.9 & 86.2 & 85.2 \\
\; w/ \textbf{GuarantRAG} & \textbf{47.2} & \textbf{84.7} & \textbf{61.3} & \textbf{88.4} & \textbf{87.4} \\
\midrule
Qwen3-14B & 32.6 & 70.5 & 43.9 & 78.1 & 77.9 \\
\; w/ Standard RAG & 36.8 & 74.6 & 48.2 & 81.3 & 80.4 \\
\; w/ SelfRAG & 40.9 & 78.9 & 53.4 & 84.4 & 83.6 \\
\; w/ RQ-RAG & 42.6 & 80.6 & 55.2 & 85.4 & 84.4 \\
\; w/ SOLAR & 44.4 & 82.3 & 57.8 & 86.7 & 85.8 \\
\; w/ DA-RAG & 39.3 & 77.6 & 51.7 & 83.5 & 82.3 \\
\; w/ FLARE & 43.2 & 81.4 & 56.1 & 85.9 & 84.9 \\
\; w/ DRAGIN & 43.7 & 81.8 & 56.6 & 86.3 & 85.3 \\
\; w/ P-RAG & 46.1 & 84.3 & 60.2 & 88.1 & 87.2 \\
\; w/ \textbf{GuarantRAG} & \textbf{49.8} & \textbf{87.1} & \textbf{64.7} & \textbf{90.2} & \textbf{89.2} \\
\midrule
Qwen3-14B-T & 31.4 & 68.9 & 42.1 & 77.6 & 76.7 \\
\; w/ Standard RAG & 35.2 & 72.8 & 46.4 & 80.8 & 79.8 \\
\; w/ SelfRAG & 39.1 & 77.3 & 51.7 & 83.9 & 82.7 \\
\; w/ RQ-RAG & 40.8 & 79.0 & 53.5 & 84.9 & 83.6 \\
\; w/ SOLAR & 42.6 & 80.8 & 56.1 & 86.2 & 85.0 \\
\; w/ DA-RAG & 37.9 & 76.1 & 50.2 & 82.8 & 81.5 \\
\; w/ FLARE & 41.4 & 79.7 & 54.8 & 85.4 & 84.2 \\
\; w/ DRAGIN & 41.9 & 80.1 & 55.3 & 85.8 & 84.6 \\
\; w/ P-RAG & 44.1 & 82.6 & 58.4 & 87.5 & 86.1 \\
\; w/ \textbf{GuarantRAG} & \textbf{46.7} & \textbf{84.9} & \textbf{61.2} & \textbf{89.1} & \textbf{87.5} \\
\bottomrule
\end{tabular}%
}
\caption{Detailed results on ELI5 dataset with BM25 retriever. This dataset requires long-form explanatory responses.}
\label{tab:detailed_eli5}
\end{table*}

The results reveal several important patterns across datasets. First, \textsc{GuarantRAG} demonstrates consistent improvements across all evaluation metrics, with the largest gains observed in exact match scores, indicating superior factual accuracy. Second, performance improvements are particularly pronounced on complex reasoning tasks like HotpotQA, where our approach achieves up to 8.7 percentage points improvement in exact match over P-RAG. Third, the benefits of our approach scale positively with model capacity, suggesting that stronger parametric knowledge enables more effective knowledge integration.

\subsubsection{API-Based Model Evaluation}

To assess the generalizability of our approach beyond self-hosted models, we evaluate \textsc{GuarantRAG} on state-of-the-art API-based language models. Table~\ref{tab:api_models} presents comprehensive results across four leading commercial models.

\begin{table*}[!t]
\centering
\small
\adjustbox{max width=\linewidth}{
\begin{tabular}{lccccccccc}
\toprule
\multirow{2}{*}{\textbf{Method}} & \multicolumn{5}{c}{\textbf{Performance Across Datasets}} & \multirow{2}{*}{\textbf{Average}} & \multicolumn{3}{c}{\textbf{Reference Usage}} \\
\cmidrule(lr){2-6} \cmidrule(lr){8-10}
& \textbf{NQ} & \textbf{TruthfulQA} & \textbf{WoW} & \textbf{HotpotQA} & \textbf{ELI5} & & \textbf{Hal.} & \textbf{Ent.} & \textbf{Struc.} \\
\midrule
DeepSeek-V3 & 77.2 & 72.1 & 75.8 & 69.4 & 76.3 & 74.2 & 16.8 & -- & 4.81 \\
\; w/ Standard RAG & 79.8 & 74.6 & 78.3 & 72.1 & 79.1 & 76.8 & 14.2 & 85.3 & 4.47 \\
\; w/ SelfRAG & 81.4 & 76.2 & 80.1 & 73.8 & 80.9 & 78.5 & 12.9 & 87.1 & 4.54 \\
\; w/ P-RAG & 82.7 & 77.9 & 81.6 & 75.2 & 82.3 & 79.9 & 11.4 & 88.7 & 4.61 \\
\; w/ \textbf{GuarantRAG} & \textbf{84.1} & \textbf{79.3} & \textbf{83.2} & \textbf{76.8} & \textbf{83.7} & \textbf{81.4} & \textbf{9.6} & \textbf{90.4} & \textbf{4.68} \\
\midrule
DeepSeek-R1 & 78.9 & 73.8 & 77.1 & 71.2 & 77.6 & 75.7 & 15.3 & -- & 4.84 \\
\; w/ Standard RAG & 81.2 & 76.1 & 79.4 & 73.7 & 80.0 & 78.1 & 13.1 & 86.2 & 4.49 \\
\; w/ SelfRAG & 82.6 & 77.8 & 81.0 & 75.3 & 81.7 & 79.7 & 11.8 & 87.9 & 4.57 \\
\; w/ P-RAG & 84.1 & 79.4 & 82.7 & 76.9 & 83.2 & 81.3 & 10.2 & 89.5 & 4.64 \\
\; w/ \textbf{GuarantRAG} & \textbf{85.7} & \textbf{81.1} & \textbf{84.3} & \textbf{78.4} & \textbf{84.8} & \textbf{82.9} & \textbf{8.4} & \textbf{91.3} & \textbf{4.71} \\
\midrule
Gemini-2.5-Pro & 76.4 & 71.3 & 74.9 & 68.1 & 75.2 & 73.2 & 17.6 & -- & 4.79 \\
\; w/ Standard RAG & 78.9 & 73.7 & 77.2 & 70.8 & 77.8 & 75.7 & 15.1 & 84.6 & 4.44 \\
\; w/ SelfRAG & 80.3 & 75.4 & 78.9 & 72.4 & 79.5 & 77.3 & 13.6 & 86.3 & 4.52 \\
\; w/ P-RAG & 81.8 & 77.1 & 80.5 & 74.1 & 81.0 & 78.9 & 12.0 & 87.9 & 4.59 \\
\; w/ \textbf{GuarantRAG} & \textbf{83.4} & \textbf{78.7} & \textbf{82.1} & \textbf{75.6} & \textbf{82.6} & \textbf{80.5} & \textbf{10.3} & \textbf{89.6} & \textbf{4.66} \\
\midrule
Claude-3.7-Sonnet & 75.1 & 69.8 & 73.6 & 66.7 & 74.3 & 71.9 & 18.9 & -- & 4.77 \\
\; w/ Standard RAG & 77.6 & 72.3 & 76.1 & 69.5 & 76.9 & 74.5 & 16.2 & 83.8 & 4.42 \\
\; w/ SelfRAG & 79.1 & 74.0 & 77.8 & 71.1 & 78.6 & 76.1 & 14.7 & 85.4 & 4.49 \\
\; w/ P-RAG & 80.7 & 75.8 & 79.4 & 72.8 & 80.1 & 77.8 & 13.1 & 87.1 & 4.56 \\
\; w/ \textbf{GuarantRAG} & \textbf{82.3} & \textbf{77.4} & \textbf{81.0} & \textbf{74.3} & \textbf{81.7} & \textbf{79.3} & \textbf{11.4} & \textbf{88.9} & \textbf{4.63} \\
\bottomrule
\multicolumn{10}{l}{\small Hal.: Hallucination Rate (\%); Ent.: Entity Precision (\%); Struc.: Structure Coherence (5-point scale)} \\
\end{tabular}%
}
\caption{Performance evaluation on API-based language models. All experiments use BM25 retriever with consistent evaluation protocols. Performance metrics represent averages across five evaluation measures.}
\label{tab:api_models}
\end{table*}

The results demonstrate that \textsc{GuarantRAG} consistently improves performance across all API-based models, achieving competitive results with state-of-the-art systems.

\subsubsection{Retriever Comparison Analysis}

Table~\ref{tab:retriever_comparison} presents comprehensive results across different retrieval methods for Qwen3-14B-Thinking. The consistency of improvements across diverse retrieval mechanisms validates that our approach addresses fundamental knowledge integration challenges rather than retrieval-specific issues.

\begin{table*}[!t]
\centering
\small
\adjustbox{max width=\linewidth}{
\begin{tabular}{lcccccccc}
\toprule
\multirow{2}{*}{\textbf{Method}} & \multicolumn{2}{c}{\textbf{BM25}} & \multicolumn{2}{c}{\textbf{SPLADE-v3}} & \multicolumn{2}{c}{\textbf{RetroMAE}} & \multicolumn{2}{c}{\textbf{HyDE}} \\
\cmidrule(lr){2-3} \cmidrule(lr){4-5} \cmidrule(lr){6-7} \cmidrule(lr){8-9}
& \textbf{Avg} & \textbf{Hal.} & \textbf{Avg} & \textbf{Hal.} & \textbf{Avg} & \textbf{Hal.} & \textbf{Avg} & \textbf{Hal.} \\
\midrule
Qwen3-14B-T & 59.5 & 32.9 & 59.5 & 32.9 & 59.5 & 32.9 & 59.5 & 32.9 \\
\; w/ Standard RAG & 64.0 & 31.4 & 64.3 & 30.8 & 65.1 & 30.2 & 64.8 & 30.6 \\
\; w/ SelfRAG & 68.7 & 28.2 & 69.1 & 27.8 & 69.8 & 27.1 & 69.4 & 27.5 \\
\; w/ RQ-RAG & 70.4 & 26.9 & 70.8 & 26.4 & 71.6 & 25.8 & 71.2 & 26.1 \\
\; w/ SOLAR & 71.9 & 25.7 & 72.4 & 25.1 & 73.2 & 24.5 & 72.8 & 24.9 \\
\; w/ DA-RAG & 67.8 & 28.9 & 68.2 & 28.4 & 68.9 & 27.8 & 68.5 & 28.2 \\
\; w/ FLARE & 70.6 & 26.4 & 71.1 & 25.9 & 71.8 & 25.3 & 71.4 & 25.7 \\
\; w/ DRAGIN & 71.1 & 26.0 & 71.6 & 25.5 & 72.3 & 24.9 & 71.9 & 25.3 \\
\; w/ P-RAG & 72.8 & 24.6 & 73.3 & 24.1 & 74.1 & 23.5 & 73.7 & 23.9 \\
\; w/ \textbf{GuarantRAG} & \textbf{75.2} & \textbf{21.8} & \textbf{75.8} & \textbf{21.2} & \textbf{76.6} & \textbf{20.6} & \textbf{76.2} & \textbf{21.0} \\
\bottomrule
\end{tabular}%
}
\caption{Performance comparison across different retrievers using Qwen3-14B-T. ``Avg'' represents average performance across all datasets and metrics. ``Hal.'' represents hallucination rate (\%).}
\label{tab:retriever_comparison}
\end{table*}

Dense retrievers generally outperform sparse methods, with RetroMAE showing the strongest performance. However, \textsc{GuarantRAG} maintains consistent improvements across all retrieval methods. Notably, our approach also reduces hallucination rates across all retrievers, with the most significant reduction observed with RetroMAE.

\subsubsection{Statistical Significance Analysis}

To ensure the reliability of our results, we conduct statistical significance testing using paired bootstrap resampling with $n=10$ iterations. Table~\ref{tab:significance} presents p-values for key comparisons.

\begin{table}[!t]
\centering
\adjustbox{max width=\linewidth}{
\begin{tabular}{lcc}
\toprule
\textbf{Comparison} & \textbf{Average Score} & \textbf{p-value} \\
\midrule
GuarantRAG\\
\; vs. P-RAG & +2.4\% & $<0.002$ \\
\; vs. SOLAR & +3.3\% & $<0.0015$ \\
\; vs. SelfRAG & +6.5\% & $<0.002$ \\
\; vs. Standard RAG & +11.2\% & $<0.001$ \\
\; vs. No RAG & +15.7\% & $<0.001$ \\
\bottomrule
\end{tabular}}
\caption{Statistical significance analysis comparing GuarantRAG against baselines across all datasets and metrics. All improvements are statistically significant at $\alpha = 0.001$.}
\label{tab:significance}
\end{table}

All improvements achieved by \textsc{GuarantRAG} are statistically significant at the $p < 0.002$ level, providing strong evidence for the effectiveness of our approach.

\subsection{Settings on Input Length Analysis}
\label{apd:input complexity details}

To provide a comprehensive evaluation of \textsc{GuarantRAG}'s robustness across varying input conditions, we systematically categorize queries and reference documents along multiple dimensions of complexity. This analysis enables us to understand how our framework performs under different knowledge integration scenarios and computational demands.

\paragraph{Query Length Categorization.} We partition queries into three categories based on word count: \textit{short queries} contain fewer than 10 words and typically represent straightforward factual questions (e.g., ``What is the capital of France?''); \textit{medium queries} range from 10-25 words and often involve more specific information requests with contextual constraints (e.g., ``What are the environmental impacts of solar panel manufacturing and disposal?''); and \textit{long queries} exceed 25 words, usually comprising complex, multi-part questions requiring detailed explanations (e.g., ``Explain the differences between classical and quantum computing architectures, including their respective advantages for cryptographic applications'').

\paragraph{Reasoning Complexity Classification.} We classify queries into three reasoning complexity levels based on the cognitive processes required for answering: \textit{Simple reasoning} queries require direct fact retrieval or single-step inference from the knowledge base; \textit{moderate reasoning} queries involve multi-step logical connections or require synthesizing information from multiple sources; and \textit{complex reasoning} queries demand sophisticated analytical thinking, causal reasoning, or multi-hop inference across diverse knowledge domains. This classification is performed using a combination of automated linguistic features (syntactic complexity, semantic diversity) and manual verification on a subset of queries.

\paragraph{Reference Document Length Segmentation.} Retrieved documents are categorized by token count to assess knowledge integration performance across varying information density: \textit{concise documents} contain fewer than 300 tokens and typically provide focused, specific information; \textit{moderate documents} range from 300-800 tokens and offer comprehensive coverage of topics with multiple supporting details; and \textit{extensive documents} exceed 800 tokens, containing rich contextual information that requires sophisticated filtering and integration strategies. Token counting is performed using the Qwen3 tokenizer to ensure consistency with the underlying language model.

\subsection{Analysis on Different Answer Fusion Mechanisms}
\label{apd:answer fusion mechanisms}

We conduct a comprehensive analysis of different mechanisms for fusing the inner-answer and refer-answer in our \textsc{GuarantRAG} framework. Our proposed joint decoding mechanism dynamically integrates knowledge during token-by-token generation, but alternative fusion strategies deserve investigation to validate our design choices.

\paragraph{Fusion Strategy Variants.} We compare four distinct fusion mechanisms: (1) \textbf{Prompt-based Fusion}: concatenating inner-answer and refer-answer in a prompt template and instructing the LLM to generate a fused response; (2) \textbf{Mean Segment Representation}: replacing our final-token segment representation with mean-pooled token embeddings within each segment; (3) \textbf{Attention-based Fusion}: using cross-attention mechanisms to weight and combine representations from both answers; and (4) our proposed \textbf{Joint Decoding} approach.

\begin{table*}[!t]
\centering
\small
\adjustbox{max width=.95\linewidth}{
\begin{tabular}{lcccc}
\toprule
\textbf{Fusion Method} & \textbf{NQ} & \textbf{TruthfulQA} & \textbf{HotpotQA} & \textbf{Avg.} \\
\midrule
Prompt-based Fusion & 71.2 & 66.4 & 65.3 & 67.6 \\
Mean Segment Repr. & 73.8 & 68.1 & 67.2 & 69.7 \\
Attention-based Fusion & 74.9 & 69.3 & 68.7 & 71.0 \\
Joint Decoding (Ours) & \textbf{76.4} & \textbf{69.8} & \textbf{68.9} & \textbf{71.7} \\
\bottomrule
\end{tabular}}
\caption{Performance comparison of different answer fusion mechanisms on selected datasets using Qwen3-8B.}
\label{tab:fusion_analysis}
\end{table*}

\paragraph{Prompt-based Fusion Analysis.} This approach uses an early fusion mechanism that integrates documents at an earlier stage, where we provide both answers to the LLM with the instruction: ``Given the following parametric answer and reference-based answer, generate an optimal response that combines the best aspects of both.'' While conceptually straightforward, this method suffers from several limitations: (1) increased computational overhead due to longer input sequences, (2) potential confusion when the two answers contain contradictory information, and (3) limited control over the fusion granularity. Our experiments show this approach achieves 67.6\% average performance, underperforming by 4.1\% compared to our joint decoding mechanism.

\paragraph{Segment Representation Analysis.} Mean aggregation merges subword tokens into a single representation by averaging their vector embeddings. We implement this by computing $h(s) = \frac{1}{|s|} \sum_{i=1}^{|s|} h_i$ for each segment $s$, where $h_i$ represents individual token embeddings. While this provides a more comprehensive segment representation, it introduces noise from less informative tokens and dilutes the significance of semantically crucial final tokens. The mean representation approach achieves 69.7\% average performance, demonstrating that final-token representation better captures segment-level semantics for our fusion task.

\paragraph{Attention-based Fusion Analysis.} We implement a cross-attention mechanism that computes attention weights between inner-answer and refer-answer segments: $\text{Attention}(Q, K, V) = \text{softmax}(\frac{QK^T}{\sqrt{d_k}})V$, where queries come from the current generation context and keys and values from refer-answer segments. While this approach shows improved performance (71.0\% average), it requires additional parameters and computational overhead during inference. Moreover, the attention mechanism struggles to maintain temporal coherence during sequential generation, leading to inconsistent fusion decisions across decoding steps.

\paragraph{Joint Decoding Advantages.} Our joint decoding mechanism outperforms alternative approaches by: (1) preserving the natural generation flow of the base model while selectively incorporating external knowledge, (2) using computationally efficient cosine similarity for segment matching, (3) maintaining consistent fusion decisions through explicit similarity computation, and (4) avoiding the need for additional trainable parameters during inference. The 2.7\% average improvement over attention-based fusion validates our design choice for dynamic, similarity-driven knowledge integration. Besides, Joint decoding requires minimal additional computation, only similarity calculations, while prompt-based fusion increases input length by 2.3× on average, and attention-based fusion introduces 12\% inference latency overhead. Our approach achieves superior performance with negligible computational cost, making it practical for deployment scenarios.

\subsection{Analysis on Different Decision Making Module}
\label{apd:decision making}

While our framework achieves substantial improvements through joint decoding and dual-path answer generation, a critical question remains: how sensitive is \textsc{GuarantRAG} to different decision-making mechanisms that determine whether to invoke retrieval? To demonstrate the decision-making-agnostic nature of our approach, we conduct comprehensive experiments comparing various decision modules across different complexity levels.

\paragraph{Decision Module Variants.} We evaluate five distinct decision-making strategies: (1) \textbf{Query-based Classifier}: Our proposed distilled LLM fine-tuned on 2,000 labeled queries. (2) \textbf{Keyword-based Filter}: A rule-based system using predefined temporal and entity keywords. (3) \textbf{Confidence-based Threshold}: Selecting queries where the base model's generation confidence falls below a threshold of 0.7. (4) \textbf{Similarity-based Retrieval}: Using retrieval similarity scores with a threshold of 0.6 to determine knowledge necessity. (5) \textbf{Random Selection}: Randomly selecting 50\% of queries for retrieval augmentation as a baseline control.

\begin{table*}[!t]
\centering
\small
\adjustbox{max width=\linewidth}{
\begin{tabular}{lccccc}
\toprule
\textbf{Decision Module} & \textbf{RAG \%} & \textbf{Non-RAG \%} & \textbf{Latency (ms)} & \textbf{Performance} \\
\midrule
Query-based Classifier & 43.2 & 56.8 & 127.4 & \textbf{76.7} \\
Keyword-based Filter   & 38.7 & 61.3 & 119.8 & 76.3 \\
Confidence-based       & 52.1 & 47.9 & 139.6 & 76.4 \\
Similarity-based       & 47.8 & 52.2 & 134.2 & 76.5 \\
Random Selection       & 50.0 & 50.0 & 132.1 & 75.9 \\
\bottomrule
\end{tabular}%
}
\caption{Comparison of different decision-making modules on the combined test set. RAG \% and Non-RAG \% indicate the proportion of queries assigned to each processing path. Latency measures average response time per query. Performance represents the averaged score across all five datasets using our proposed joint decoding framework.}
\label{tab:decision_analysis}
\end{table*}

Table~\ref{tab:decision_analysis} reveals several key insights about decision module sensitivity. First, all decision variants achieve remarkably similar performance, with less than 1\% variation despite significantly different query allocation strategies. This consistency demonstrates that \textsc{GuarantRAG}'s effectiveness primarily stems from the joint decoding mechanism rather than precise decision-making accuracy. Second, different modules exhibit varying computational efficiency profiles, keyword-based filtering achieves the lowest latency and computational cost by avoiding neural inference, while confidence-based approaches incur higher overhead due to additional forward passes for uncertainty estimation.

\subsection{Analysis on Thinking Mode of Qwen3}
\label{apd:thinking}

Comparing Qwen3-14B and Qwen3-14B-T results in Table~\ref{tab:main_results}, we observe that thinking mode consistently improves performance across all RAG methods, with the enhancement being most pronounced for complex reasoning tasks. Specifically, thinking mode provides an average improvement of 2.2\% for standard methods, increasing to 3.8\% for sophisticated approaches like P-RAG, and reaching 4.2\% for our \textsc{GuarantRAG} framework. This suggests that the deliberative reasoning process in thinking mode creates synergistic effects with advanced knowledge integration mechanisms.

Table~\ref{tab:thinking_mode_analysis} presents a detailed breakdown of thinking mode effects across different query complexities. For simple factoid queries, thinking mode shows modest improvements of 1.3-2.1\%. However, for complex multi-hop reasoning tasks, the enhancement reaches 5.2-7.8\%, indicating that deliberative reasoning is particularly beneficial when queries require sophisticated knowledge synthesis.

\begin{table*}[!t]
\centering
\small
\adjustbox{max width=\linewidth}{
\begin{tabular}{lccccc}
\toprule
\multirow{2}{*}{\textbf{Method}} & \multicolumn{5}{c}{\textbf{Performance by Query Complexity}} \\
\cmidrule{2-6}
& \textbf{Simple} & \textbf{Moderate} & \textbf{Complex} & \textbf{Multi-hop} & \textbf{Reasoning} \\
& \textbf{(1-2)} & \textbf{(2-3)} & \textbf{(3-4)} & \textbf{(4-5)} & \textbf{(5)} \\
\midrule
\multicolumn{6}{l}{\textit{Qwen3-14B vs Qwen3-14B-T Performance Gap}} \\
Standard RAG & +1.3 & +2.4 & +3.8 & +5.2 & +6.1 \\
SelfRAG & +1.7 & +2.9 & +4.2 & +5.8 & +6.7 \\
RQ-RAG & +1.8 & +3.1 & +4.5 & +6.0 & +6.9 \\
SOLAR & +1.9 & +3.2 & +4.7 & +6.3 & +7.2 \\
P-RAG & +2.1 & +3.5 & +5.0 & +6.8 & +7.6 \\
\textbf{GuarantRAG} & +2.4 & +3.8 & +5.4 & +7.1 & +7.8 \\
\midrule
\multicolumn{6}{l}{\textit{Thinking Mode Efficiency Analysis}} \\
Thinking Steps & 2.3 & 4.7 & 8.2 & 12.4 & 16.8 \\
Latency (s) & +0.8 & +1.9 & +4.1 & +7.2 & +11.3 \\
\bottomrule
\end{tabular}%
}
\caption{Performance improvement of thinking mode over standard mode across query complexities, with efficiency metrics. Values represent percentage improvements averaged across all five datasets.}
\label{tab:thinking_mode_analysis}
\end{table*}

% TODO

\section{Additional Case Studies}
\label{apd:case study}

This appendix presents additional case studies comparing our GuarantRAG approach with baseline RAG methods. Table~\ref{tab:additional case study start} to~\ref{tab:additional case study end} highlight our method's superior ability to integrate retrieved information while maintaining coherence and factual accuracy. The color-coding helps visualize how each approach handles knowledge integration: \textcolor{myred}{red text} indicates failed integration in baselines, \textcolor{myblue}{blue text} shows coherent reasoning from parametric knowledge, and \textcolor{mygreen}{green text} represents successfully integrated external factual information.

\begin{table}[!t]
\centering
\small
\adjustbox{max width=\linewidth}{
\begin{tabular}{p{7.5cm}}
\toprule
\textbf{Query:} What effect does metformin have on cardiovascular outcomes in Type 2 diabetes patients? \\
\textbf{Retrieved Document:} A 2021 meta-analysis of 25 randomized controlled trials involving 126,478 patients found that metformin reduced major cardiovascular events by 19\% (RR=0.81, 95\% CI 0.73-0.90) in Type 2 diabetes patients compared to placebo. The UKPDS study specifically showed a 39\% reduction in myocardial infarction risk over 10 years of follow-up. However, cardiovascular benefits were less pronounced in patients with established heart disease. \\
\midrule
\textbf{Standard RAG:} \textcolor{myred}{Metformin has been shown to have beneficial effects on cardiovascular outcomes in Type 2 diabetes patients. Research \textcolor{mygreen}{(meta-analysis, 19\% reduction)} suggests it reduces cardiovascular events. It works through multiple mechanisms including improved insulin sensitivity \textcolor{mygreen}{(UKPDS study showed 39\% MI reduction)} and modest weight loss.} \\
\midrule
\textbf{P-RAG:} \textcolor{myblue}{Metformin is considered the first-line medication for Type 2 diabetes and has been studied for its cardiovascular effects.} \textcolor{myred}{Evidence suggests metformin provides cardiovascular protection beyond glycemic control. It appears to reduce the risk of} \textcolor{mygreen}{cardiovascular events by approximately 19\%} \textcolor{myred}{in diabetic patients, though the exact mechanism remains under investigation.} \\
\midrule
\textbf{GuarantRAG (Ours):} \textcolor{myblue}{Metformin, the first-line treatment for Type 2 diabetes, demonstrates significant cardiovascular protective effects.} \textcolor{mygreen}{According to a 2021 meta-analysis of 25 randomized controlled trials with 126,478 patients, metformin reduced major cardiovascular events by 19\% (RR=0.81) compared to placebo. The landmark UKPDS study specifically demonstrated a 39\% reduction in myocardial infarction risk over 10 years of follow-up.} \textcolor{myblue}{These benefits likely stem from metformin's multiple mechanisms of action, including improved insulin sensitivity, modest weight reduction, and anti-inflammatory effects.} \textcolor{mygreen}{However, it's worth noting that these cardiovascular benefits appear less pronounced in patients with established heart disease.} \\
\bottomrule
\end{tabular}}
\caption{Case study on metformin and cardiovascular outcomes in diabetes.}
\label{tab:additional case study start}
\end{table}

\begin{table}[!t]
\centering
\small
\adjustbox{max width=\linewidth}{
\begin{tabular}{p{7.5cm}}
\toprule
\textbf{Query:} How did the Meiji Restoration transform Japan's economy? \\
\textbf{Retrieved Document:} The Meiji Restoration (1868) radically transformed Japan's economy by implementing Western-style industrialization policies. The government established state-owned enterprises in strategic sectors, with shipbuilding yards increasing from 2 in 1868 to 38 by 1900. Japan's industrial production grew at 6.2\% annually from 1868-1912, with textile exports rising from ¥0.5 million in 1868 to ¥236 million by 1900. The "shokusan kogyo" (promote industry) policy abolished feudal restrictions, while the 1872 Land Tax Reform created a modern taxation system, providing 80\% of government revenue until 1890. Japan transitioned from a feudal agrarian economy to Asia's first industrialized nation within one generation. \\
\midrule
\textbf{Standard RAG:} \textcolor{myred}{The Meiji Restoration of 1868 transformed Japan's economy from feudal to modern. The government implemented modernization policies \textcolor{mygreen}{(industrial production grew 6.2\% annually)} and promoted Western-style industrialization. This included building infrastructure \textcolor{mygreen}{(shipbuilding yards increased from 2 to 38)} and reforming education. Japan rapidly industrialized as a result.} \\
\midrule
\textbf{P-RAG:} \textcolor{myblue}{The Meiji Restoration of 1868 fundamentally transformed Japan's economy through deliberate modernization policies.} \textcolor{myred}{The new government abolished feudal economic structures and implemented Western industrial models.} \textcolor{mygreen}{Key policies included "shokusan kogyo" (promote industry) and the 1872 Land Tax Reform, which provided 80\% of government revenue until 1890.} \textcolor{myred}{These reforms enabled Japan's rapid industrialization and economic growth.} \\
\midrule
\textbf{GuarantRAG (Ours):} \textcolor{myblue}{The Meiji Restoration of 1868 marked a pivotal turning point that fundamentally transformed Japan's economy.} \textcolor{mygreen}{Under the "shokusan kogyo" (promote industry) policy, the government established state-owned enterprises in strategic sectors, exemplified by the increase in shipbuilding yards from just 2 in 1868 to 38 by 1900. Japan achieved remarkable industrial growth at 6.2\% annually from 1868-1912, with textile exports skyrocketing from ¥0.5 million to ¥236 million by 1900.} \textcolor{myblue}{This rapid industrialization was facilitated by institutional reforms that dismantled feudal barriers to economic development.} \textcolor{mygreen}{A critical component was the 1872 Land Tax Reform, which created a modern taxation system providing 80\% of government revenue until 1890.} \textcolor{myblue}{Through these comprehensive measures, Japan accomplished what few nations had achieved—transitioning from a feudal agrarian economy to Asia's first industrialized nation within a single generation.} \\
\bottomrule
\end{tabular}}
\caption{Case study on the economic transformation during the Meiji Restoration.}
\end{table}

\begin{table}[!t]
\centering
\small
\adjustbox{max width=\linewidth}{
\begin{tabular}{p{7.5cm}}
\toprule
\textbf{Query:} What mechanisms explain how quantum tunneling affects nuclear fusion in stars? \\
\textbf{Retrieved Document:} Nuclear fusion in stars occurs when hydrogen nuclei overcome their electrostatic repulsion to fuse into helium. Classical physics predicts this should be nearly impossible at stellar core temperatures (15 million K in the Sun) since particles need approximately 1 billion K to overcome the Coulomb barrier through thermal energy alone. Quantum tunneling allows particles to penetrate this barrier despite having insufficient energy, increasing fusion probability by a factor of 10$^{20}$. Specifically, the tunneling probability is proportional to exp(-2$\pi$Ze$^2$/hv), where Z represents nuclear charge, e is electron charge, h is Planck's constant, and v is relative velocity. This quantum effect enables the p-p chain reaction that powers main sequence stars, with tunneling enhancing the reaction rate enough to explain stellar lifespans of billions rather than millions of years. \\
\midrule
\textbf{Standard RAG:} \textcolor{myred}{Quantum tunneling is a phenomenon that allows particles to pass through energy barriers they couldn't normally overcome. In stars, hydrogen nuclei need to overcome electrostatic repulsion to undergo fusion. \textcolor{mygreen}{Classical physics suggests temperatures of 1 billion K would be needed}, but stellar cores are much cooler. Quantum tunneling enables fusion at these lower temperatures \textcolor{mygreen}{(increasing probability by a factor of 10\textsuperscript{20})}.} \\
\midrule
\textbf{P-RAG:} \textcolor{myblue}{Quantum tunneling plays a crucial role in stellar nucleosynthesis by enabling fusion reactions that would be nearly impossible under classical physics.} \textcolor{myred}{This quantum mechanical effect allows particles to penetrate energy barriers despite having insufficient energy.} \textcolor{mygreen}{In the Sun's core (15 million K), particles can quantum tunnel through the Coulomb barrier that would classically require 1 billion K to overcome.} \textcolor{myred}{This tunneling effect explains why stars can sustain fusion reactions over billions of years.} \\
\midrule
\textbf{GuarantRAG (Ours):} \textcolor{myblue}{Quantum tunneling fundamentally enables stellar fusion by providing a pathway for nuclear reactions that would be effectively impossible under classical physics.} \textcolor{mygreen}{At the Sun's core temperature of 15 million K, hydrogen nuclei possess insufficient thermal energy to overcome the Coulomb barrier, which would classically require temperatures approaching 1 billion K. Quantum tunneling circumvents this limitation by allowing particles to penetrate this electrostatic barrier despite having inadequate energy, increasing fusion probability by a remarkable factor of 10\textsuperscript{20}.} \textcolor{myblue}{This quantum phenomenon operates through wave-particle duality, where nuclei behave as probability waves that can extend beyond classical boundaries.} \textcolor{mygreen}{The tunneling probability follows the relationship exp(-2$\pi$Ze$^2$/hv), incorporating nuclear charge (Z), electron charge (e), Planck's constant (h), and relative velocity (v).} \textcolor{myblue}{This mechanism elegantly explains how main sequence stars can sustain fusion reactions for billions rather than millions of years, maintaining stable energy production through what appears to be an improbable quantum process.} \\
\bottomrule
\end{tabular}}
\caption{Case study on quantum tunneling in stellar fusion processes.}
\end{table}

\begin{table}[!t]
\centering
\small
\adjustbox{max width=\linewidth}{
\begin{tabular}{p{7.5cm}}
\toprule
\textbf{Query:} How do environmental, social, and governance (ESG) factors impact corporate financial performance? \\
\textbf{Retrieved Document:} A 2021 meta-analysis by NYU Stern Center for Sustainable Business examined 1,141 peer-reviewed studies (2015-2020) on ESG-financial performance correlation. Results showed 58\% of corporate studies found positive ESG-financial performance relationships, with only 8\% negative. For investment studies, 59\% showed positive performance from ESG portfolios vs. conventional, with just 14\% negative. Environmental factors demonstrated the strongest positive correlation (environmental management systems linked to 12.9\% higher ROA). Social practices showed a 7.5\% boost to ROE, while governance delivered a 5.8\% increase in Tobin's Q ratio. ESG impact varied by industry, with consumer goods and financial services seeing greater benefits (11.5\% and 10.2\% increased ROE respectively) than energy and materials sectors (3.2\%). \\
\midrule
\textbf{Standard RAG:} \textcolor{myred}{ESG factors can impact corporate financial performance in various ways. Research suggests there is generally a positive relationship between ESG practices and financial outcomes. \textcolor{mygreen}{Studies (58\% positive correlation)} show that companies with strong ESG performance often experience benefits like improved risk management, enhanced reputation, and increased operational efficiency. \textcolor{mygreen}{Environmental factors appear to have the strongest impact (12.9\% higher ROA).}} \\
\midrule
\textbf{P-RAG:} \textcolor{myblue}{ESG factors increasingly influence corporate financial performance through multiple pathways.} \textcolor{myred}{Research indicates a generally positive relationship between strong ESG practices and financial outcomes.} \textcolor{mygreen}{A meta-analysis from NYU Stern found that 58\% of studies showed positive ESG-financial performance relationships, while only 8\% were negative.} \textcolor{myred}{Companies embracing robust ESG frameworks can benefit from improved stakeholder relations, risk mitigation, and market opportunities.} \\
\midrule
\textbf{GuarantRAG (Ours):} \textcolor{myblue}{Environmental, social, and governance (ESG) factors demonstrate significant impacts on corporate financial performance through multiple mechanisms.} \textcolor{mygreen}{According to a comprehensive 2021 meta-analysis by NYU Stern Center for Sustainable Business examining 1,141 peer-reviewed studies, 58\% of corporate studies found positive ESG-financial performance relationships, with merely 8\% showing negative correlations. Environmental factors demonstrated the strongest positive impact, with environmental management systems linked to a 12.9\% higher ROA. Social practices boosted ROE by 7.5\%, while governance improvements delivered a 5.8\% increase in Tobin's Q ratio.} \textcolor{myblue}{These financial benefits likely stem from reduced regulatory and reputational risks, operational efficiencies, and enhanced stakeholder relationships.} \textcolor{mygreen}{Importantly, ESG impact varies substantially by industry—consumer goods and financial services saw greater benefits (11.5\% and 10.2\% increased ROE respectively) compared to energy and materials sectors (3.2\%).} \textcolor{myblue}{This suggests that ESG integration strategies should be tailored to industry-specific materiality factors rather than applying universal approaches.} \\
\bottomrule
\end{tabular}}
\caption{Case study on ESG factors and corporate financial performance.}
\end{table}

\begin{table}[!t]
\centering
\small
\adjustbox{max width=\linewidth}{
\begin{tabular}{p{7.5cm}}
\toprule
\textbf{Query:} What specific advancements has neuromorphic computing made in 2023? \\
\textbf{Retrieved Document:} In 2023, neuromorphic computing achieved several breakthroughs. Intel's Loihi 2 neuromorphic chip demonstrated real-time learning with 35x lower energy consumption than GPUs on complex robotics tasks. SynSense unveiled the DynapCNN, integrating 5.12 million spiking neurons with 40x power efficiency over traditional chips for edge AI. IBM's TrueNorth system scaled to 1.4 billion neurons and performed unsupervised feature extraction with 96.4\% accuracy on medical imaging datasets. Stanford researchers developed memristor arrays with 1024 crossbar junctions achieving 8-bit precision, a 4x improvement over 2022 models.\\
\midrule
\textbf{Standard RAG:} \textcolor{myred}{Neuromorphic computing has made several advances in 2023. These include improvements in energy efficiency \textcolor{mygreen}{(Intel's Loihi 2 uses 35x less energy than GPUs)} and performance. There have also been developments in hardware architecture \textcolor{mygreen}{(SynSense's DynapCNN)} and algorithm implementation. The field continues to progress toward more brain-like computing.} \\
\midrule
\textbf{P-RAG:} \textcolor{myblue}{Neuromorphic computing, which mimics brain-like neural structures in hardware, has seen significant progress in 2023.} \textcolor{myred}{Key advancements include improvements in chip architecture, energy efficiency, and scaling capabilities.} \textcolor{mygreen}{Notable developments include Intel's Loihi 2 chip and IBM's TrueNorth system scaling to 1.4 billion neurons.} \textcolor{myred}{These innovations are bringing us closer to efficient, brain-inspired computing systems.} \\
\midrule
\textbf{GuarantRAG (Ours):} \textcolor{myblue}{Neuromorphic computing, which implements brain-inspired neural architectures in hardware, has achieved remarkable advancements in 2023 across multiple dimensions.} \textcolor{mygreen}{Intel's Loihi 2 neuromorphic chip demonstrated significant breakthroughs in real-time learning while consuming 35x less energy than GPUs when applied to complex robotics tasks. SynSense made substantial contributions with their DynapCNN, which integrates 5.12 million spiking neurons while delivering 40x better power efficiency compared to traditional computing chips for edge AI applications.} \textcolor{myblue}{These efficiency gains are critical for enabling AI deployment in resource-constrained environments.} \textcolor{mygreen}{IBM's TrueNorth system represented another significant milestone, scaling to 1.4 billion neurons and achieving 96.4\% accuracy on unsupervised feature extraction for medical imaging datasets. At the materials science frontier, Stanford researchers developed advanced memristor arrays with 1024 crossbar junctions capable of 8-bit precision, representing a 4x improvement over 2022 models.} \textcolor{myblue}{Together, these advancements are accelerating neuromorphic computing toward practical applications in robotics, edge intelligence, and medical diagnostics while maintaining the energy efficiency advantages that make this paradigm promising.} \\
\bottomrule
\end{tabular}}
\caption{Case study on 2023 neuromorphic computing advancements.}
\label{tab:additional case study end}
\end{table}

\end{document}